\documentclass[10pt,journal,compsoc]{IEEEtran}
\usepackage{epsfig}
\usepackage{graphicx}
\usepackage{amsmath}
\usepackage{amssymb}
\usepackage{tabularx}
\usepackage{subcaption}
\usepackage{xcolor}
\usepackage{threeparttable}
\usepackage{algorithm}
\usepackage{algorithmic}
\usepackage{enumitem}
\usepackage{multirow}
\usepackage{lipsum}
\usepackage[pagebackref=false,breaklinks=true,colorlinks,bookmarks=false]{hyperref}
\ifCLASSOPTIONcompsoc
  \usepackage[nocompress]{cite}
\else
  \usepackage{cite}
\fi

\ifCLASSINFOpdf
\else
\fi

\usepackage{xspace}
\makeatletter
\DeclareRobustCommand\onedot{\futurelet\@let@token\@onedot}
\def\@onedot{\ifx\@let@token.\else.\null\fi\xspace}

\def\eg{\emph{e.g}\onedot} 
\def\ie{\emph{i.e}\onedot} 
 
 \def\vs{\emph{vs}\onedot}

\newcolumntype{P}[1]{>{\centering\arraybackslash}p{#1}}
\newcommand{\blue}[1]{\textcolor{black}{#1}}

\newcommand{\black}[1]{\textcolor{black}{#1}}
\begin{document}
%
\title{Deep Learning for Person Re-identification: \\A Survey and Outlook  }
%
%
\author{Mang Ye, Jianbing Shen,~\IEEEmembership{Senior Member,~IEEE}, Gaojie Lin, Tao Xiang\\
Ling Shao and Steven C. H. Hoi,~\IEEEmembership{Fellow, IEEE}
\IEEEcompsocitemizethanks{
\IEEEcompsocthanksitem M. Ye is with the School of Computer Science, Wuhan University, China and Inception Institute of Artificial Intelligence, UAE.
\IEEEcompsocthanksitem J. Shen and L. Shao are with the Inception Institute of Artificial Intelligence, UAE. \protect
E-mail:\{mangye16, shenjianbingcg\}@gmail.com
\IEEEcompsocthanksitem G. Lin is with the School of Computer Science, Beijing Institute of Technology, China.
\IEEEcompsocthanksitem T. Xiang is with the Centre for Vision Speech and Signal Processing, University of Surrey, UK.  \protect Email:  t.xiang@surrey.ac.uk
\IEEEcompsocthanksitem S. C. H. Hoi is with the Singapore Management
University, and Salesforce Research Asia, Singapore.  \protect Email: stevenhoi@gmail.com
}
}

%
%

\markboth{IEEE Transactions on Pattern Analysis and Machine Intelligence}%
{Shell \MakeLowercase{\textit{et al.}}: Bare Demo of IEEEtran.cls for Computer Society Journals}

\IEEEtitleabstractindextext{%
\begin{abstract}
Person re-identification (Re-ID) aims at retrieving a person of interest across multiple non-overlapping cameras. With the advancement of deep neural networks and increasing demand of intelligent video surveillance, it has gained significantly increased interest in the computer vision community. By dissecting the involved components in developing a person Re-ID system, we categorize it into the closed-world and open-world settings. The widely studied closed-world setting is usually applied under various research-oriented assumptions, and has achieved inspiring success using deep learning techniques on a number of datasets. We first conduct a comprehensive overview with in-depth analysis for closed-world person Re-ID from three different perspectives, including deep feature representation learning, deep metric learning and ranking optimization. With the performance saturation under closed-world setting, the research focus for person Re-ID has recently shifted to the open-world setting, facing more challenging issues. This setting is closer to practical applications under specific scenarios. We summarize the open-world Re-ID in terms of five different aspects. By analyzing the advantages of existing methods, we design a powerful AGW baseline, achieving state-of-the-art or at least comparable performance on twelve datasets for FOUR different Re-ID tasks. Meanwhile, we introduce a new evaluation metric (mINP) for person Re-ID, indicating the cost for finding all the correct matches, which provides an additional criteria to evaluate the Re-ID system for real applications. Finally, some important yet under-investigated open issues are discussed.
\end{abstract}

\begin{IEEEkeywords}
Person Re-Identification, Pedestrian Retrieval, Literature Survey, Evaluation Metric, Deep Learning
\end{IEEEkeywords}}

\maketitle

\IEEEdisplaynontitleabstractindextext

\IEEEpeerreviewmaketitle

\IEEEraisesectionheading{\section{Introduction}\label{sec:intro}}

\blue{\IEEEPARstart{P}{erson} re-identification (Re-ID) has been widely studied as a specific person retrieval problem across non-overlapping cameras \cite{pami17craft,survey16}. Given a query person-of-interest, the goal of Re-ID is to determine whether this person has appeared in another place at a distinct time captured by a different camera, or even the same camera at a different time instant \cite{cvpr06reid}. The query person can be represented by an image \cite{arxiv18right,iccv15zheng,cvprw19aggregate}, a video sequence \cite{eccv14video,eccv16mars}, and even a text description \cite{icmr15,iccv17nlp}. Due to the urgent demand of public safety and increasing number of surveillance cameras, person Re-ID is imperative in intelligent surveillance systems with significant research impact and practical importance.}

\blue{Re-ID is a challenging task due to the presence of different viewpoints \cite{iccv15viewpoint,avss14viewpoint}, varying low-image resolutions \cite{iccv15lowres,cvpr18multires}, illumination changes \cite{mm19illumination}, unconstrained poses \cite{cvpr16pose,cvpr17spindle,cvpr18pose}, occlusions \cite{cvpr18advocclus,cvpr19occlusion},
heterogeneous modalities \cite{iccv17cross,iccv17nlp}, complex camera environments, background clutter \cite{cvpr18mask}, unreliable bounding box generations, etc. These result in varying variations and uncertainty. In addition, for practical model deployment, the dynamic updated camera network \cite{cviu17camera,eccv16temporal}, large scale gallery with efficient retrieval \cite{tip17rank}, group uncertainty \cite{pami15openworld}, significant domain shift \cite{icip15active}, unseen testing scenarios \cite{cvpr19general}, incremental model updating \cite{eccv14camera} and changing cloths \cite{pami19cloth} also greatly increase the difficulties. These challenges lead that Re-ID is still unsolved problem.
Early research efforts mainly focus on the hand-crafted feature construction with body structures \cite{eccv08elf,cvpr10sdalf,eccv14yang,cvpr15lomo,cvpr16gog} or distance metric learning \cite{cvpr12kissme,cvpr11prdc,eccv14kernelmetric,eccv12metric,iccv15liao,pami18unsupervised}.
With the advancement of deep learning, person Re-ID has achieved inspiring performance on the widely used benchmarks \cite{iccv15zheng,iccv17duke,cvpr14cuhk,cvpr18msmt}. However, there is still a large gap between the research-oriented scenarios and practical applications \cite{tcsvt19survey}. This motivates us to conduct a comprehensive survey, develop a powerful baseline for different Re-ID tasks and discuss several future directions.}

\black{Though some surveys have also summarized the deep learning techniques \cite{survey16,survey19neuro,survey18arxiv}, our survey makes three major differences: 1) We provide an in-depth and comprehensive analysis of existing deep learning methods by discussing their advantages and limitations, analyzing the state-of-the-arts. This provides insights for future algorithm design and new topic exploration. 2) We design a new powerful baseline (AGW: Attention Generalized mean pooling with Weighted triplet loss) and a new evaluation metric (mINP: mean Inverse Negative Penalty) for future developments. AGW achieves state-of-the-art performance on twelve datasets for four different Re-ID tasks. mINP provides a supplement metric to existing CMC/mAP, indicating the cost to find all the correct matches. 3) We make an attempt to discuss several important research directions with under-investigated open issues to narrow the gap between the closed-world and open-world applications, taking a step towards real-world Re-ID system design. }
\begin{figure*}[t]
  \centering
  \includegraphics[height =3cm,width = 17.6cm]{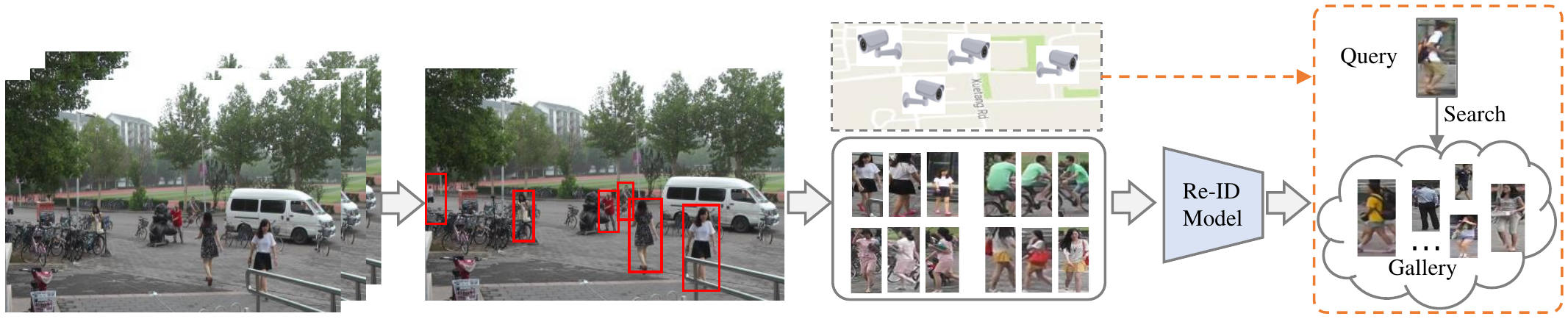}\\
  \vspace{-2mm}
  \caption{The flow of designing a practical person Re-ID system, including five main steps: 1) \textit{Raw Data Collection}, (2) \textit{Bounding Box Generation}, 3) \textit{Training Data Annotation}, 4) \textit{Model Training} and 5) \textit{Pedestrian Retrieval}.}\label{fig:flow}
  \vspace{-4mm}
\end{figure*}

Unless otherwise specified, person Re-ID in this survey refers to the pedestrian retrieval problem across multiple surveillance cameras, from a computer vision perspective. Generally, building a person Re-ID system for a specific scenario requires five main steps (as shown in Fig.~\ref{fig:flow}):
\vspace{-1mm}
\begin{enumerate}
\item  \label{item:step1}Step 1: \textit{Raw Data Collection}: Obtaining raw video data from surveillance cameras is the primary requirement of practical video investigation. These cameras are usually located in different places under varying environments \cite{prl13survey}. Most likely, this raw data contains a large amount of complex and noisy background clutter.
  \item \label{item:step2} Step 2: \textit{Bounding Box Generation}: Extracting the bounding boxes which contain the person images from the raw video data. Generally, it is impossible to manually crop all the person images in large-scale applications. The bounding boxes are usually obtained by the person detection \cite{pami09detecsurvey,cvpr09detectbencmark} or tracking algorithms \cite{cvpr17persontrack,cvpr18multicam}.
  \item \label{item:step3}Step 3: \textit{Training Data Annotation}: Annotating the cross-camera labels. Training data annotation is usually indispensable for discriminative Re-ID model learning due to the large cross-camera variations. In the existence of large domain shift \cite{iccv13domain}, we often need to annotate the training data in every new scenario.
  \item \label{item:step4} Step 4: \textit{Model Training}: Training a discriminative and robust Re-ID model with the previous annotated person images/videos. This step is the core for developing a Re-ID system and it is also the most widely studied paradigm in the literature. Extensive models have been developed to handle the various challenges, concentrating on feature representation learning \cite{cvpr16dgd,cvpr17prw}, distance metric learning \cite{icpr14deepmetric,arxiv17triplet} or their combinations.
  \item \label{item:step5}Step 5: \textit{Pedestrian Retrieval}: The testing phase conducts the pedestrian retrieval. Given a person-of-interest (query) and a gallery set, we extract the feature representations using the Re-ID model learned in previous stage. A retrieved ranking list is obtained by sorting the calculated query-to-gallery similarity. Some methods have also investigated the ranking optimization to improve the retrieval performance \cite{cvpr17rerank,tmm16rank}.
\end{enumerate}

\begin{table}[t]
\caption{\small{Closed-world \vs Open-world Person Re-ID.}}\label{tab:openclose}
\vspace{-2mm}
\begin{tabular}{l|l}
  \hline
  Closed-world~(Section \ref{sec:close})& Open-world~(Section \ref{sec:open})\\\hline
    $\checkmark$ Single-modality Data & Heterogeneous Data (\S~\ref{sec:hetro})\\
  $\checkmark$ Bounding Boxes Generation & Raw Images/Videos (\S~\ref{sec:end2end})\\
  $\checkmark$ Sufficient Annotated Data & Unavailable/Limited Labels (\S~\ref{sec:unsuper})\\
 $\checkmark$ Correct Annotation  & Noisy Annotation (\S~\ref{sec:noise})\\
 $\checkmark$ Query Exists in Gallery & Open-set (\S~\ref{sec:openset})\\\hline
\end{tabular}
\vspace{-3mm}
\end{table}

According to the five steps mentioned above, we categorize existing Re-ID methods into two main trends: \textit{closed-world} and \textit{open-world} settings, as summarized in Table \ref{tab:openclose}. A step-by-step comparison is in the following five aspects:
 \vspace{-1mm}
\begin{enumerate}
\item  \black{\textit{Single-modality \vs Heterogeneous Data}: For the raw data collection in Step \ref{item:step1}, all the persons are represented by images/videos captured by single-modality visible cameras in the closed-world setting \cite{iccv15zheng,eccv16mars,iccv17duke,cvpr14cuhk,eccv08elf,cvpr18msmt}. However, in practical open-world applications, we might also need to process heterogeneous data, which are infrared images \cite{iccv17cross,sensors17}, sketches \cite{pr19sketch}, depth images \cite{tip17rgbd}, or even text descriptions \cite{cvpr17nlp}. This motivates the heterogeneous Re-ID in \S~\ref{sec:hetro}.}
  \item \textit{Bounding Box Generation \vs Raw Images/Videos }: For the bounding box generation in Step \ref{item:step2}, the closed-world person Re-ID usually performs the training and testing based on the generated bounding boxes, where the bounding boxes mainly contain the person appearance information. In contrast, some practical open-world applications require end-to-end person search from the raw images or videos \cite{cvpr17prw,cvpr17joint}. This leads to another open-world topic, \ie, end-to-end person search in \S~\ref{sec:end2end}.
  \item \textit{Sufficient Annotated Data \vs Unavailable/Limited Labels}: For the training data annotation in Step \ref{item:step3}, the closed-world person Re-ID usually assumes that we have enough annotated training data for supervised Re-ID model training. However, label annotation for each camera pair in every new environment is time consuming and labor intensive, incurring high costs. In open-world scenarios, we might not have enough annotated data (\ie, limited labels) \cite{cvpr14semi} or even without any label information \cite{cvpr13saliency}. This inspires the discussion of the unsupervised and semi-supervised Re-ID in \S~\ref{sec:unsuper}.
  \item \black{\textit{Correct Annotation  \vs Noisy Annotation}: For Step \ref{item:step4}, existing closed-world person Re-ID systems usually assume that all the annotations are correct, with clean labels. However, annotation noise is usually unavoidable due to annotation error (\ie, label noise) or imperfect detection/tracking results (\ie, sample noise, partial Re-ID \cite{cvpr19partial}). This leads to the analysis of noise-robust person Re-ID under different noise types in \S~\ref{sec:noise}.}
  \item \textit{Query Exists in Gallery \vs Open-set}: In the pedestrian retrieval stage (Step \ref{item:step5}), most existing closed-world person Re-ID works assume that the query must occur in the gallery set by calculating the CMC \cite{iccv07cmc} and mAP \cite{iccv15zheng}. However, in many scenarios, the query person may not appear in the gallery set \cite{icip16openset,tip18open}, or we need to perform the verification rather than retrieval \cite{pami15openworld}. This brings us to the open-set person Re-ID in \S~\ref{sec:openset}.
\end{enumerate}

This survey first introduces the widely studied person Re-ID under closed-world settings in \S~\ref{sec:close}. A detailed review on the datasets and the state-of-the-arts are conducted in \S~\ref{sec:data}. We then introduce the open-world person Re-ID in \S~\ref{sec:open}. An outlook for future Re-ID is presented in \S~\ref{sec:future}, including a new evaluation metric (\S~\ref{sec:newmetric}), a new powerful AGW baseline (\S~\ref{sec:newbase}). We discuss several under-investigated open issues for future study (\S~\ref{sec:openissue}). Conclusions will be drawn in \S~\ref{sec:conclusion}. A structure overview is shown in the supplementary.

\begin{figure*}[t]
  \centering
  \includegraphics[height=3cm]{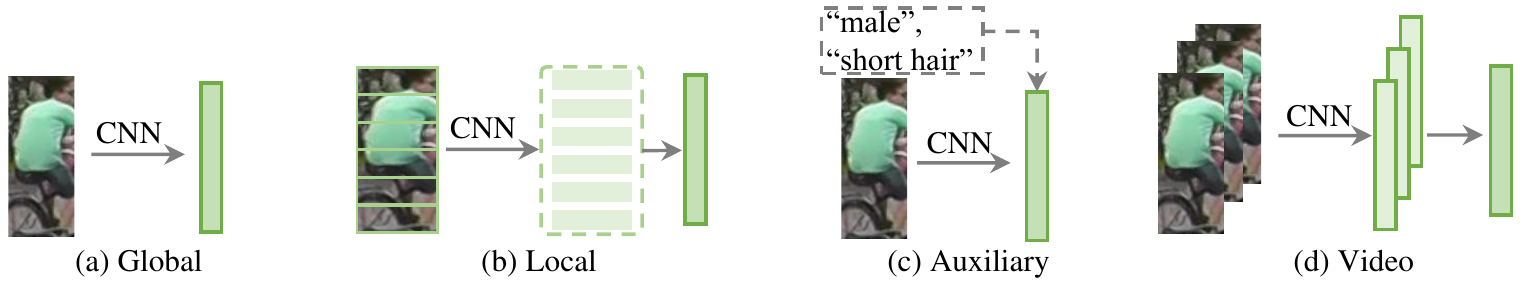}\\
  \vspace{-2mm}
  \caption{Four different feature learning strategies. a) Global Feature, learning a global representation for each person image in \S~\ref{sec:featglobal}; b) Local Feature, learning part-aggregated local features in \S~\ref{sec:featlocal}; c) Auxiliary Feature, learning the feature representation using auxiliary information, \eg, attributes \cite{eccv16attribute,arxiv17att} in \S~\ref{sec:feataux} and d) Video Feature , learning the video representation using multiple image frames and temporal information \cite{iccv15des,tip18videotemp} in \S~\ref{sec:featvideo}. }\label{fig:featclass}
  \vspace{-2mm}
\end{figure*}
\section{Closed-world Person Re-Identification}\label{sec:close}
This section provides an overview for closed-world person Re-ID. As discussed in \S~\ref{sec:intro}, this setting usually has the following assumptions: 1) person appearances are captured by single-modality visible cameras, either by image or video; 2) The persons are represented by bounding boxes, where most of the bounding box area belongs the same identity; 3) The training has enough annotated training data for supervised discriminative Re-ID model learning; 4) The annotations are generally correct; 5) The query person must appear in the gallery set. Typically, a standard closed-world Re-ID system contains three main components: \textit{Feature Representation Learning} (\S~\ref{sec:feat}), which focuses on developing the feature construction strategies; \textit{Deep Metric Learning} (\S~\ref{sec:metric}), which aims at designing the training objectives with different loss functions or sampling strategies; and \textit{Ranking Optimization} (\S~\ref{sec:rank}), which concentrates on optimizing the retrieved ranking list. An overview of the datasets and state-of-the-arts with in-depth analysis is provided in \S~\ref{sec:datasota}.

\subsection{Feature Representation Learning}\label{sec:feat}
We firstly discuss the feature learning strategies in closed-world person Re-ID. There are four main categories (as shown in Fig.~\ref{fig:featclass}): a) Global Feature (\S~\ref{sec:featglobal}), it extracts a global feature representation vector for each person image without additional annotation cues \cite{cvpr17prw}; b) Local Feature (\S~\ref{sec:featlocal}), it aggregates part-level local features to formulate a combined representation for each person image \cite{cvpr17partalign,tip19part,eccv18pcb}; c) Auxiliary Feature (\S~\ref{sec:feataux}), it improves the feature representation learning using auxiliary information, \eg, attributes \cite{eccv16attribute,arxiv17att,icpr16attributes}, GAN generated images \cite{iccv17duke}, etc. d) Video Feature  (\S~\ref{sec:featvideo}), it learns video representation for video-based Re-ID \cite{eccv14video} using multiple image frames and temporal information \cite{iccv15des,tip18videotemp}. We also review several specific architecture designs for person Re-ID in \S~\ref{sec:featarchi}.

\subsubsection{Global Feature Representation Learning}\label{sec:featglobal}
Global feature representation learning extracts a global feature vector for each person image, as shown in Fig.~\ref{fig:featclass}(a). Since deep neural networks are originally applied in image classification \cite{arxvi14vgg,cvpr16resnet}, global feature learning is the primary choice when integrating advanced deep learning techniques into the person Re-ID field in early years.

\black{To capture the fine-grained cues in global feature learning,
A joint learning framework consisting of a single-image representation (SIR) and cross-image representation (CIR) is developed in \cite{cvpr16jointsir}, trained with triplet loss using specific sub-networks.
The widely-used ID-discriminative Embedding (IDE) model \cite{cvpr17prw} constructs the training process as a multi-class classification problem by treating each identity as a distinct class. It is now widely used in Re-ID community \cite{iccv17duke,iccv17svd,cvpr17rerank,eccv18pcb,eccv18race}.
Qian \textit{et al.} \cite{iccv17multiscale} develop a multi-scale deep representation learning model to capture discriminative cues at different scales.}

\black{\textbf{Attention Information.} Attention schemes have been widely studied in literature to enhance representation learning \cite{pr19attention}. 1) \textit{Group 1: Attention within the person image.} Typical strategies include the pixel level attention \cite{cvpr18hann} and the channel-wise feature response re-weighting \cite{cvpr18hann,eccv18mancs,cvpr18kronecker,cvpr18cascadedconv}, or background suppressing \cite{cvpr18mask}. The spatial information is integrated in \cite{iccv19selfattention}. 2) \textit{Group 2: attention across multiple person images.} A context-aware attentive feature learning method is proposed in \cite{cvpr18dualatt}, incorporating both an intra-sequence and inter-sequence attention for pair-wise feature alignment and refinement. The attention consistency property is added in \cite{cvpr19siameseattend,iccv19consistatt}. Group similarity \cite{cvpr18groupcrf,iccv19sft} is another popular approach to leverage the cross-image attention, which involves multiple images for local and global similarity modeling.
The first group mainly enhances the robustness against misalignment/imperfect detection, and the second improves the feature learning by mining the relations across multiple images. }


\subsubsection{Local Feature Representation Learning}\label{sec:featlocal}
\black{It learns part/region aggregated features, making it robust against misalignment \cite{eccv16lstmpart,eccv18pcb}. The body parts are either automatically generated by human parsing/pose estimation (Group 1) or roughly horizontal division  (Group 2).}


\black{With automatic body part detection, the popular solution is to combine the full body representation and local part features \cite{eccv18partalign,iccv17partalign}. Specifically, the multi-channel aggregation \cite{cvpr16part}, multi-scale context-aware convolutions \cite{cvpr17contextpart}, multi-stage feature decomposition \cite{cvpr17spindle} and bilinear-pooling \cite{eccv18partalign} are designed to improve the local feature learning. Rather than feature level fusion, the part-level similarity combination is also studied in  \cite{iccv17partalign}. Another popular solution is to enhance the robustness against background clutter, using the pose-driven matching \cite{iccv17pose}, pose-guided part attention module \cite{cvpr18partatt}, semantically part alignment \cite{cvpr19denselyalign,iccv19dualpart}.  }

\black{For horizontal-divided region features, multiple part-level classifiers are learned in Part-based Convolutional Baseline (PCB) \cite{eccv18pcb}, which now serves as a strong part feature learning baseline in the current state-of-the-art \cite{cvpr19viewpoint,cvpr19exemplarmemory,cvpr19general}. To capture the relations across multiple body parts, the Siamese Long Short-Term Memory (LSTM) architecture \cite{eccv16lstmpart}, second-order non-local attention \cite{iccv19secondnonlocal}, Interaction-and-Aggregation (IA) \cite{cvpr19interact} are designed to reinforce the feature learning.}


\black{The first group uses human parsing techniques to obtain semantically meaningful body parts, which provides well-align part features. However, they require an additional pose detector and are prone to noisy pose detections \cite{eccv18pcb}. The second group uses a uniform partition to obtain the horizontal stripe parts, which is more flexible, but it is sensitive to heavy occlusions and large background clutter.}


\subsubsection{Auxiliary Feature Representation Learning}\label{sec:feataux}
Auxiliary feature representation learning usually requires additional annotated information (\eg, semantic attributes \cite{eccv16attribute}) or generated/augmented training samples to reinforce the feature representation \cite{iccv17duke,cvpr18advocclus}.

\textbf{Semantic Attributes}. A joint identity and attribute learning baseline is introduced in \cite{arxiv17att}. Su \textit{et al.} \cite{eccv16attribute} propose a deep attribute learning framework by incorporating the predicted semantic attribute information, enhancing the generalizability and robustness of the feature representation in a semi-supervised learning manner. Both the semantic attributes and the attention scheme are incorporated to improve part feature learning \cite{cvpr19aanet}. Semantic attributes are also adopted in \cite{cvpr19videoatt} for video Re-ID feature representation learning. They  are also leveraged as the auxiliary supervision information in unsupervised learning \cite{cvpr18att}.


\black{\textbf{Viewpoint Information.}
The viewpoint information is also leveraged to enhance the feature representation learning \cite{cvpr18multfac,iccv19view}. Multi-Level Factorisation Net (MLFN) \cite{cvpr18multfac} also tries to learn the identity-discriminative and view-invariant feature representations at multiple semantic levels. Liu \textit{et al.} \cite{iccv19view} extract a combination of view-generic and view-specific learning. An angular regularization is incorporated in \cite{aaai20val} in the viewpoint-aware feature learning.}

\textbf{Domain Information.}
A Domain Guided Dropout (DGD) algorithm \cite{cvpr16dgd} is designed to adaptively mine the domain-sharable and domain-specific neurons for multi-domain deep feature representation learning. Treating each camera as a distinct domain, Lin \textit{et al.} \cite{cvpr17consistent} propose a multi-camera consistent matching constraint to obtain a globally optimal representation in a deep learning framework. Similarly, the camera view information or the detected camera location is also applied in \cite{cvpr18pose} to improve the feature representation with camera-specific information modeling.

\black{\textbf{GAN Generation.}
This section discusses the use of GAN generated images as the auxiliary information. Zheng \textit{et al.} \cite{iccv17duke} start the first attempt to apply the GAN technique for person Re-ID. It improves the supervised feature representation learning with the generated person images. Pose constraints are incorporated in \cite{cvpr18posetrans} to improve the quality of the generated person images, generating the person images with new pose variants. A pose-normalized image generation approach is designed in \cite{eccv18posenorm}, which enhances the robustness against pose variations. Camera style information \cite{cvpr18camera} is also integrated in the image generation process to address the cross camera variations. A joint discriminative and generative learning model \cite{cvpr19joint} separately learns the appearance and structure codes to improve the image generation quality. Using the GAN generated images is also a widely used approach in unsupervised domain adaptation Re-ID  \cite{cvpr18spgan,iccv19crosspose}, approximating the target distribution.}

\black{\textbf{Data Augmentation.}
For Re-ID, custom operations are random resize, cropping and horizontal flip \cite{arxiv19trick}. Besides, adversarially occluded samples \cite{cvpr18advocclus} are generated to augment the variation of training data. A similar random erasing strategy is proposed in \cite{arxiv17randomerase}, adding random noise to the input images. A batch DropBlock \cite{iccv19dropblock} randomly drops a region block in the feature map to reinforce the attentive feature learning. Bak \textit{et al.} \cite{eccv18bak} generate the virtual humans rendered under different illumination conditions.
These methods enrich the supervision with the augmented samples, improving the generalizability on the testing set.}


\subsubsection{Video Feature Representation Learning}\label{sec:featvideo}
Video-based Re-ID is another popular topic \cite{prid2011}, where each person is represented by a video sequence with multiple frames. Due to the rich appearance and temporal information, it has gained increasing interest in the Re-ID community. This also brings in additional challenges in video feature representation learning with multiple images.


\black{The primary challenge is to accurately capture the temporal information. A recurrent neural network architecture is designed for video-based person Re-ID \cite{cvpr16video}, which jointly optimizes the final recurrent layer for temporal information propagation and the temporal pooling layer. A weighted scheme for spatial and temporal streams is developed in \cite{iccv17videotwo}.
Yan \textit{et al.} \cite{eccv16aggregate} present a progressive/sequential fusion framework to aggregate the frame-level human region representations.
Semantic attributes are also adopted in \cite{cvpr19videoatt} for video Re-ID with feature disentangling and frame re-weighting. Jointly aggregating the frame-level feature and spatio-temporal appearance information is crucial for video representation learning \cite{cvpr17forest,iccv17joint,iccv19coseg}.}

\black{Another major challenge is the unavoidable outlier tracking frames within the videos.
Informative frames are selected in a joint Spatial and Temporal Attention Pooling Network (ASTPN) \cite{iccv17joint}, and the contextual information is integrated in \cite{cvpr17forest}. A co-segmentation inspired attention model \cite{iccv19coseg} detects salient features across multiple video frames with mutual consensus estimation. A diversity regularization \cite{cvpr18diversityatt} is employed to mine multiple discriminative body parts in each video sequence.
An affine hull is adopted to handle the outlier frames within the video sequence \cite{eccv18race}. An interesting work \cite{cvpr19occlusion} utilizes the multiple video frames to auto-complete occluded regions. These works demonstrate that handling the noisy frames can greatly improve the video representation learning.}

\black{It is also challenging to handle the varying lengths of video sequences, Chen \textit{et al.} \cite{cvpr18snippet} divide the long video sequences into multiple short snippets, aggregating the top-ranked snippets to learn a compact embedding. A clip-level learning strategy \cite{aaai19sta} exploits both spatial and temporal dimensional attention cues to produce a robust clip-level representation. Both the short- and long-term relations \cite{iccv19longshort} are integrated in a self-attention scheme.}


\subsubsection{Architecture Design}\label{sec:featarchi}
Framing person Re-ID as a specific pedestrian retrieval problem, most existing works adopt the network architectures \cite{arxvi14vgg,cvpr16resnet} designed for image classification as the backbone. Some works have tried to modify the backbone architecture to achieve better Re-ID features. For the widely used ResNet50 backbone \cite{cvpr16resnet}, the important modifications include changing the last convolutional stripe/size to 1 \cite{eccv18pcb}, employing adaptive average pooling in the last pooling layer \cite{eccv18pcb}, and adding bottleneck layer with batch normalization after the pooling layer \cite{iccv17svd}.

\black{Accuracy is the major concern for specific Re-ID network architecture design to improve the accuracy, Li \textit{et al.} \cite{cvpr14cuhk} start the first attempt by designing a filter pairing neural
network (FPNN), which jointly handles misalignment and occlusions with part discriminative information mining.
Wang \textit{et al.} \cite{cvpr18cascadedconv} propose a BraidNet with a specially designed WConv layer and Channel Scaling layer. The WConv layer extracts the difference information of two images to enhance the robustness against misalignments and Channel Scaling layer optimizes the scaling factor of each input channel.
A Multi-Level Factorisation Net (MLFN) \cite{cvpr18multfac} contains multiple stacked blocks to model various latent factors at a specific level, and the factors are dynamically selected to formulate the final representation. An efficient fully convolutional Siamese network \cite{cvpr18efficient} with convolution similarity module is developed to optimize multi-level similarity measurement. The similarity is efficiently captured and optimized by using the depth-wise convolution.}

\black{Efficiency is another important factor for Re-ID architecture design. An efficient small scale network, namely Omni-Scale Network (OSNet) \cite{iccv19osnet}, is designed by incorporating the point-wise and depth-wise convolutions.  To achieve multi-scale feature learning, a
residual block composed of multiple convolutional streams is introduced.}

With the increasing interest in auto-machine learning, an Auto-ReID \cite{iccv19autoreid} model is proposed. Auto-ReID provides an efficient and effective automated neural architecture design based on a set of basic architecture components, using a part-aware module to capture the discriminative local Re-ID features. This provides a potential research direction in exploring powerful domain-specific architectures.

%
%

\begin{figure}[t]
  \centering
  \includegraphics[width = 8.8cm]{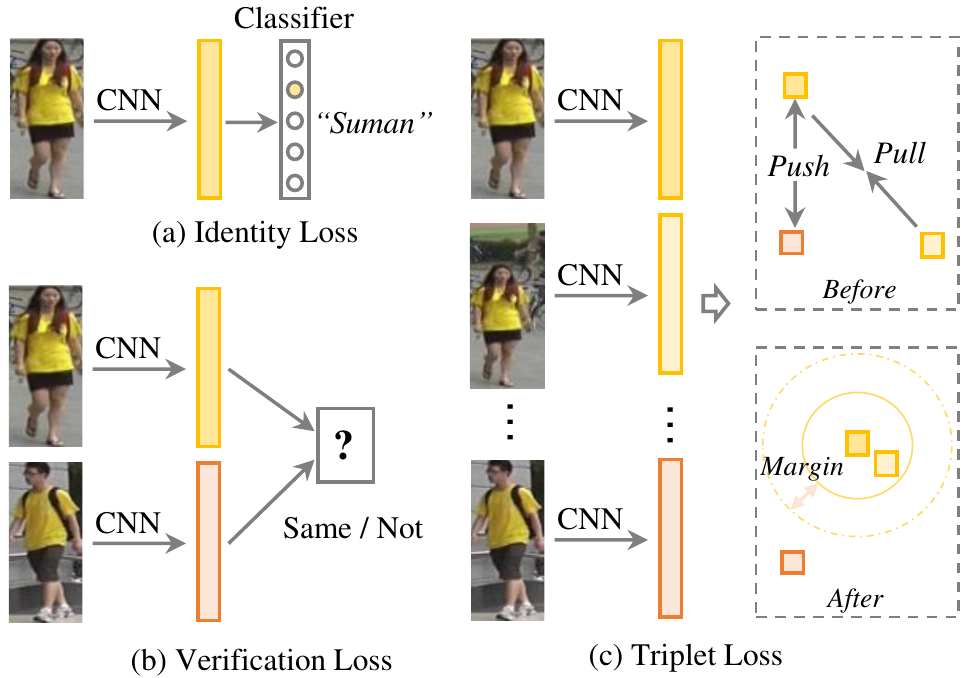}\\
  \vspace{-2mm}
  \caption{Three kinds of widely used loss functions in the literature. (a) Identity Loss  \cite{iccv17svd,iccv17duke,cvpr18camera,cvpr18background} ; (b) Verification Loss \cite{cvpr18groupcrf,arxiv16deep} and (c) Triplet Loss \cite{arxiv17triplet,cvpr18mask,cvpr18multires}. Many works employ their combinations \cite{cvpr18efficient,eccv18mancs,tifs19vtreid,arxiv16deep}. }\label{fig:metric}
  \vspace{-2mm}
\end{figure}
\subsection{Deep Metric Learning}\label{sec:metric}
Metric learning has been extensively studied before the deep learning era by learning a Mahalanobis distance function \cite{cvpr11prdc,cvpr12kissme} or projection matrix \cite{iccv15liao}. The role of metric learning has been replaced by the loss function designs to guide the feature representation learning. We will first review the widely used loss functions in \S~\ref{sec:metricloss} and then summarize the training strategies with specific sampling designs \S~\ref{sec:metrictrain}.
\subsubsection{Loss Function Design}\label{sec:metricloss}
This survey only focuses on the loss functions designed for deep learning \cite{icpr14deepmetric}. An overview of the distance metric learning designed for hand-crafted systems can be found in \cite{survey16,survey16metric}. There are three widely studied loss functions with their variants in the literature for person Re-ID, including the identity loss, verification loss and triplet loss. An illustration of three loss functions is shown in Fig.~\ref{fig:metric}.

\textbf{Identity Loss.}
It treats the training process of person Re-ID as an image classification problem \cite{cvpr17prw}, \ie, each identity is a distinct class. In the testing phase, the output of the pooling layer or embedding layer is adopted as the feature extractor. Given an input image $x_i$ with label $y_i$, the predicted probability of $x_i$ being recognized as class $y_i$ is encoded with a softmax function, represented by $p(y_i|x_i)$. The identity loss is then computed by the cross-entropy
\begin{equation}\label{eq:identityloss}
\mathcal{L}_{id}  = -\frac{1}{n} \sum\nolimits_{i = 1}^{n} {\log(p(y_i|x_i) )},
\end{equation}
where $n$ represents the number of training samples within each batch. The identity loss has been widely used in existing methods \cite{iccv17svd,iccv17duke,cvpr18spgan,cvpr18advocclus,cvpr18wu,cvpr18camera,cvpr18background,cvpr19exemplarmemory,cvpr19siameseattend,iccv19sft}. Generally, it is easy to train and automatically mine the hard samples during the training process, as demonstrated in \cite{cvpr19uel}. Several works have also investigated the softmax variants \cite{wacv18cosine}, such as the sphere loss in \cite{spherereid19} and AM softmax in \cite{iccv19sft}. Another simple yet effective strategy, \ie, label smoothing \cite{iccv17duke,arxiv19trick}, is generally integrated into the standard softmax cross-entropy loss. Its basic idea is to avoid the model fitting to over-confident annotated labels, improving the generalizability \cite{arxiv19labelsmooth}.

\textbf{Verification Loss.} It optimizes the pairwise relationship, either with a contrastive loss \cite{eccv16lstmpart,cvpr18spgan} or binary verification loss \cite{arxiv16deep,cvpr14cuhk}. The contrastive loss improves the relative pairwise distance comparison, formulated by
 \begin{equation}\label{eq:contrastive}
\mathcal{L}_{con}  = (1- \delta_{ij})\{\max(0, \rho - d_{ij})\}^2 + \delta_{ij} d_{ij}^2,
\end{equation}
where $d_{ij}$ represents the Euclidean distance between the embedding features of two input samples $x_i$ and $x_j$. $\delta_{ij}$ is a binary label indicator ($\delta_{ij} =1$ when $x_i$ and $x_j$ belong to the same identity, and $\delta_{ij} =0$, otherwise). $\rho$ is a margin parameter. There are several variants, \eg, the pairwise comparison with ranking SVM in \cite{cvpr16jointsir}.

Binary verification \cite{arxiv16deep,cvpr14cuhk} discriminates the positive and negative of a input image pair. Generally, a differential feature $f_{ij}$ is obtained by $f_{ij} = (f_{j}- f_{j})^2$ \cite{arxiv16deep}, where $f_{i}$ and $f_{j}$ are the embedding features of two samples $x_i$ and $x_j$. The verification network classifies the differential feature into positive or negative. We use $p(\delta_{ij}|f_{ij})$ to represent the probability of an input pair ($x_i$ and $x_j$) being recognized as $\delta_{ij}$ (0 or 1). The verification loss with cross-entropy is
\begin{equation}\label{eq:veriloss}
\mathcal{L}_{veri}(i,j) = - {\delta_{ij} \log(p(\delta_{ij}|f_{ij})) } - (1-\delta_{ij})\log(1-p(\delta_{ij}|f_{ij})).
\end{equation}

The verification is often combined with the identity loss to improve the performance \cite{cvpr18groupcrf,arxiv16deep,cvpr18spgan,eccv16lstmpart}.

\textbf{Triplet loss.} It treats the Re-ID model training process as a retrieval ranking problem. The basic idea is that the distance between the positive pair should be smaller than the negative pair by a pre-defined margin \cite{arxiv17triplet}. Typically, a triplet contains one anchor sample $x_i$, one positive sample $x_j$ with the same identity, and one negative sample $x_k$ from a different identity. The triplet loss with a margin parameter is represented by
\begin{equation}\label{eq:triloss}
\mathcal{L}_{tri}(i,j,k) = \max (\rho + d_{ij} - d_{ik}, 0),
\end{equation}
where $d(\cdot)$ measures the Euclidean distance between two samples.
\black{The large proportion of easy triplets will dominate the training process if we directly optimize above loss function, resulting in limited discriminability. To alleviate this issue, various informative triplet mining methods have been designed \cite{arxiv17triplet,cvpr18mask,cvpr18multires,eccv18partalign}. The basic idea is to select the informative triplets \cite{arxiv17triplet,eccv16deepmetric}.
Specifically, a moderate positive mining with a weight constraint is introduced in \cite{eccv16deepmetric}, which directly optimizes the feature difference. Hermans \textit{et al.} \cite{arxiv17triplet} demonstrate that the online hardest positive and negative mining within each training batch is beneficial for discriminative Re-ID model learning.
Some methods also studied the point to set similarity strategy for informative triplet mining \cite{cvpr17point2set,eccv18hardaware}. This enhances robustness against the outlier samples with a soft hard-mining scheme.}

To further enrich the triplet supervision, a quadruplet deep network is developed in \cite{cvpr17quadloss}, where each quadruplet contains one anchor sample, one positive sample and two mined negative samples. The quadruplets are formulated with a margin-based online hard negative mining. Optimizing the quadruplet relationship results in smaller intra-class variation and larger inter-class variation.

The combination of triplet loss and identity loss is one of the most popular solutions for deep Re-ID model learning \cite{cvpr18efficient,cvpr18posetrans,eccv18mancs,tifs19vtreid,cvpr19denselyalign,cvpr19general,cvpr19unpatch,iccv19dualpart,iccv19humanloop,iccv19consistatt,iccv19selfattention}. These two components are mutually beneficial for discriminative feature representation learning.

\textbf{OIM loss.}
In addition to the above three kinds of loss functions, an Online Instance Matching (OIM) loss \cite{cvpr17joint} is designed with a memory bank scheme. A memory bank $\{v_k,k=1,2,\cdots,c\}$ contains the stored instance features, where $c$ denotes the class number. The OIM loss is then formulated by
\begin{equation}\label{eq:oimloss}
\mathcal{L}_{oim}  = -\frac{1}{n} \sum\nolimits_{i = 1}^{n} {\log \frac{\exp (v_i^T f_i / \tau)}{\sum\nolimits_{k=1}^{c}\exp (v_k^T f_i / \tau)} },
\end{equation}
where $v_i$ represents the corresponding stored memory feature for class $y_i$, and $\tau$ is a temperature parameter that controls the similarity space \cite{cvpr19uel}. $v_i^T f_i$ measures the online instance matching score. The comparison with a memorized feature set of unlabelled identities is further included to calculate the denominator \cite{cvpr17joint}, handling the large instance number of non-targeted identities. This memory scheme is also adopted in unsupervised domain adaptive Re-ID \cite{cvpr19exemplarmemory}.


\subsubsection{Training strategy}\label{sec:metrictrain}
\black{The batch sampling strategy plays an important role in discriminative Re-ID model learning. It is challenging since the number of annotated training images for each identity varies significantly \cite{iccv15zheng}. Meanwhile, the severely imbalanced positive and negative sample pairs increases additional difficulty for the training strategy design \cite{iccv15liao}.}

\black{The most commonly used training strategy for handling the imbalanced issue is identity sampling \cite{arxiv17triplet,arxiv19trick}. For each training batch, a certain number of identities are randomly selected, and then several images are sampled from each selected identity. This batch sampling strategy guarantees the informative positive and negative mining.}

\black{To handle the imbalance issue between the positive and negative, adaptive sampling is the popular approach to adjust the contribution of positive and negative samples, such as Sample Rate Learning (SRL) \cite{cvpr18cascadedconv}, curriculum sampling \cite{eccv18mancs}. Another approach is sample re-weighting, using the sample distribution \cite{eccv18mancs} or similarity difference \cite{cvpr18multicam} to adjust the sample weight.
An efficient reference constraint is designed in \cite{cvpr18reference} to transform the pairwise/triplet similarity to a sample-to-reference similarity, addressing the imbalance issue and enhancing the discriminability, which is also robust to outliers.}

\black{To adaptively combine multiple loss functions, a multi-loss dynamic training strategy \cite{cvpr19multi} adaptively reweights the identity loss and triplet loss, extracting appropriate component shared between them. This multi-loss training strategy leads to consistent performance gain.}
\begin{figure}[t]
  \centering
  \includegraphics[width = 8cm]{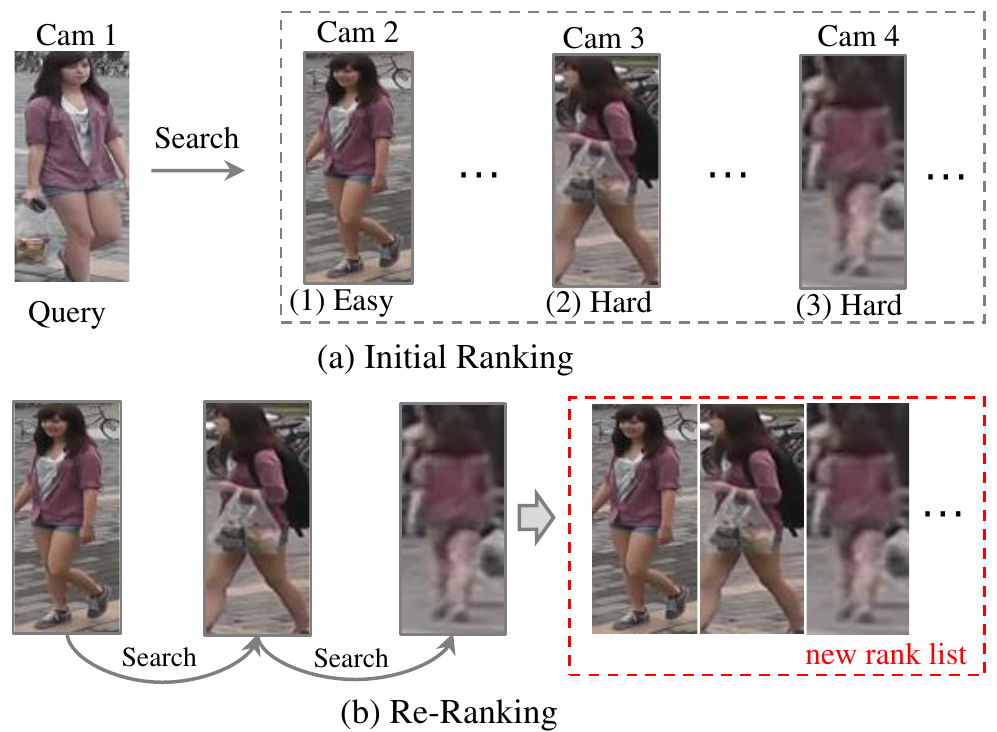}\\
  \vspace{-2mm}
  \caption{An illustration of re-ranking in person Re-ID. Given a query example, an initial rank list is retrieved, where the hard matches are ranked in the bottom. Using the top-ranked easy positive match (1) as query to search in the gallery, we can get the hard match (2) and (3) with similarity propagation in the gallery set.}\label{fig:rerank}
  \vspace{-2mm}
\end{figure}
\subsection{Ranking Optimization}\label{sec:rank}
Ranking optimization plays a crucial role in improving the retrieval performance in the testing stage. Given an initial ranking list, it optimizes the ranking order, either by automatic gallery-to-gallery similarity mining \cite{mm15rank,cvpr17rerank} or human interaction \cite{iccv13pop,eccv16humanloop}. Rank/Metric fusion \cite{cvpr15metricensemble,cvpr19rerank} is another popular approach for improving the ranking performance with multiple ranking list inputs.

\subsubsection{Re-ranking}\label{sec:rankre}
\black{The basic idea of re-ranking is to utilize the gallery-to-gallery similarity to optimize the initial ranking list, as shown in Fig.~\ref{fig:rerank}. The top-ranked similarity pulling and bottom-ranked dissimilarity pushing is proposed in \cite{mm15rank}. The widely-used $k$-reciprocal reranking \cite{cvpr17rerank} mines the contextual information. Similar idea for contextual information modeling is applied in \cite{tip17rank}. Bai \textit{et al.} \cite{cvpr17bai} utilize the geometric structure of the underlying manifold. An expanded cross neighborhood re-ranking method \cite{cvpr18pose} is introduced by integrating the cross neighborhood distance.
A local blurring re-ranking \cite{iccv19sft} employs the clustering structure to improve neighborhood similarity measurement.}

\textbf{Query Adaptive.}
Considering the query difference, some methods have designed the query adaptive retrieval strategy to replace the uniform searching engine to improve the performance \cite{accv14queryadap,iccv17onlinelocal}. Andy \textit{et al.} \cite{accv14queryadap} propose a query adaptive re-ranking method using locality preserving projections.
An efficient online local metric adaptation method is presented in \cite{iccv17onlinelocal}, which learns a strictly local metric with mined negative samples for each probe.

\textbf{Human Interaction.}
It involves using human feedback to optimize the ranking list \cite{iccv13pop}. This provides reliable supervision during the re-ranking process. A hybrid human-computer incremental learning model is presented in \cite{eccv16humanloop}, which cumulatively learns from human feedback, improving the Re-ID ranking performance on-the-fly.

\subsubsection{Rank Fusion}\label{sec:rankfusion}
Rank fusion exploits multiple ranking lists obtained with different methods to improve the retrieval performance \cite{tmm16rank}. Zheng \textit{et al.} \cite{cvpr15latefusion} propose a query adaptive late fusion method on top of a ``L" shaped observation to fuse methods. A rank aggregation method by employing the similarity and dissimilarity is developed in \cite{tmm16rank}.
The rank fusion process in person Re-ID is formulated as a consensus-based decision problem with graph theory \cite{iccv17shape}, mapping the similarity scores obtained by multiple algorithms into a graph with path searching.
An Unified Ensemble Diffusion (UED) \cite{cvpr19rerank} is recently designed for metric fusion. UED maintains the advantages of three existing fusion algorithms, optimized by a new objective function and derivation. The metric ensemble learning is also studied in \cite{cvpr15metricensemble}.
%

\setlength{\tabcolsep}{3.75pt}
\begin{table}[t]
\scriptsize
\centering
\caption{Statistics of some commonly used datasets for closed-world person Re-ID. ``both" means that it contains both hand-cropped and detected bounding boxes. ``C\&M" means both CMC and mAP are evaluated.}
\vspace{-2mm}
\begin{tabular}{l|ccccccc}
\hline
 & \multicolumn{7}{c}{\textit{Image datasets}}\\
Dataset& Time& \#ID & \#image & \#cam. & Label & Res. & Eval.\\
\hline
VIPeR& 2007 & 632 & 1,264 &2 & hand& fixed&CMC\\
iLIDS& 2009  & 119 & 476 &2 & hand& vary&CMC\\
GRID&  2009 & 250 &1,275 &8 & hand& vary&CMC\\
PRID2011&  2011 & 200 &1,134 &2 & hand& fixed&CMC\\
CUHK01& 2012  & 971 & 3,884&2 & hand& fixed&CMC\\
CUHK02& 2013 & 1,816 & 7,264 &10 & hand & fixed&CMC\\
CUHK03& 2014 & 1,467 & 13,164 &2 & both & vary&CMC\\
Market-1501& 2015  & 1,501 & 32,668 &6 & both &fixed &C\&M\\
DukeMTMC & 2017 & 1,404 & 36,411 & 8 & both & fixed & C\&M\\
Airport & 2017 & 9,651 & 39,902 & 6 & auto & fixed & C\&M \\
MSMT17 & 2018 & 4,101 & 126,441& 15 & auto & vary & C\&M \\
\hline
 & \multicolumn{7}{c}{\textit{Video datasets}}\\
 Dataset& time& \#ID &\#track(\#bbox) & \#cam. & label & Res. & Eval\\
\hline
PRID-2011&  2011 & 200 &400 (40k)&2 & hand&fixed&CMC\\
iLIDS-VID& 2014  & 300&600 (44k) &2 & hand&vary&CMC\\
MARS& 2016  & 1261&20,715 (1M) &6 & auto & fixed&C\&M\\
Duke-Video & 2018 & 1,812 & 4,832 (-) & 8 & auto & fixed & C\&M\\
Duke-Tracklet & 2018 & 1,788 & 12,647 (-) & 8 & auto & C\&M \\
LPW & 2018 &  	2,731 & 7,694(590K) & 4 & auto & fixed & C\&M \\
LS-VID & 2019 & 3,772 & 14,943 (3M) & 15 & auto & fixed & C\&M \\\hline
\end{tabular}
\label{tab:dataset}
\vspace{-3mm}
\end{table}

\begin{figure*}[t]
  \centering
    \begin{minipage}[t]{0.25\textwidth}
        \centering
        \vspace{-0.2cm}
        \includegraphics[height = 4.3cm, width=4.5cm]{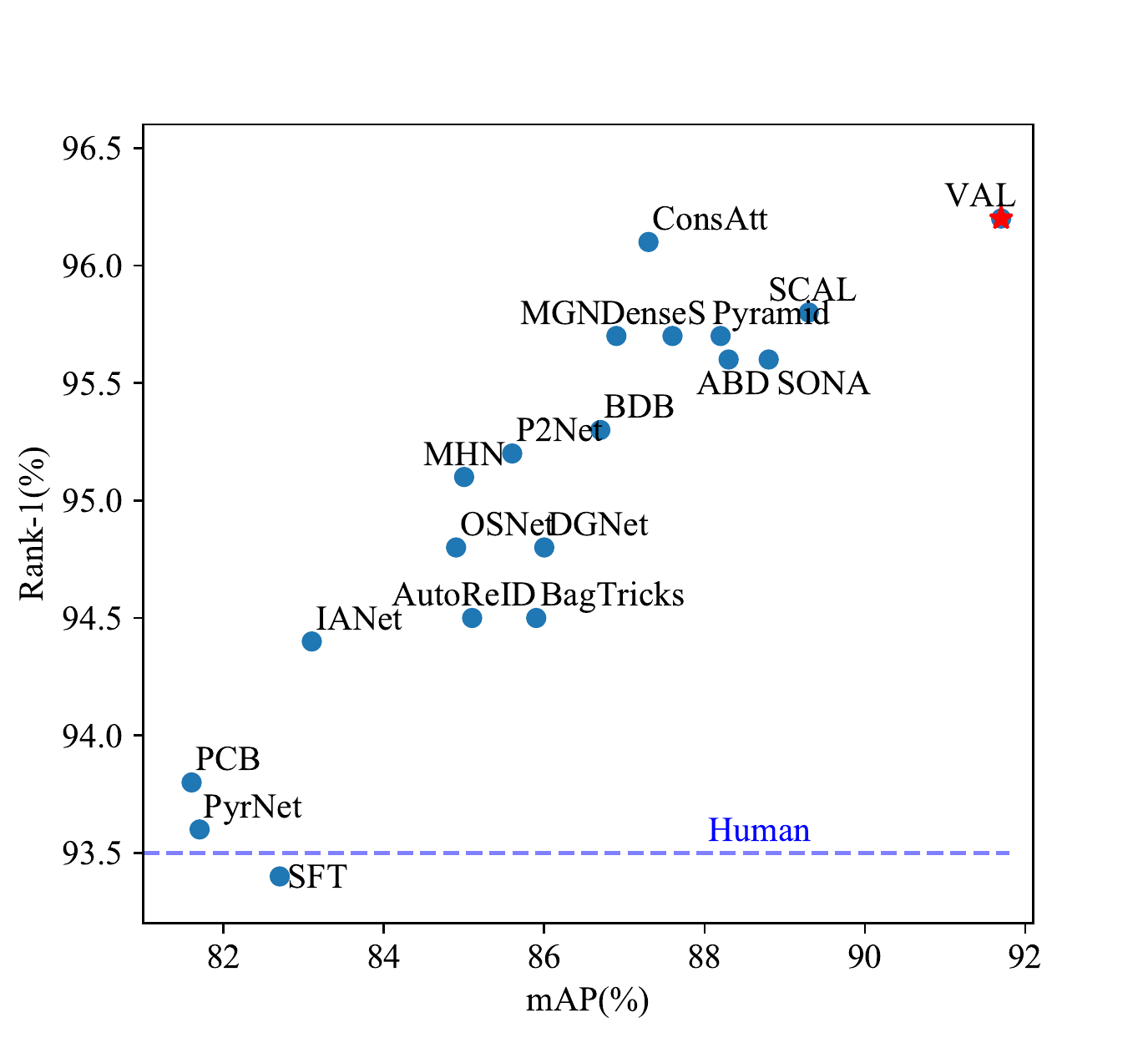}
        \vspace{-0.2cm}
        \centerline{\scriptsize{(a) SOTA on Market-1501 \cite{iccv15zheng} }}\medskip
        \vspace{-2mm}
    \end{minipage}%
    \begin{minipage}[t]{0.25\textwidth}
        \centering
        \vspace{-0.2cm}
        \includegraphics[height = 4.3cm,width=4.5cm]{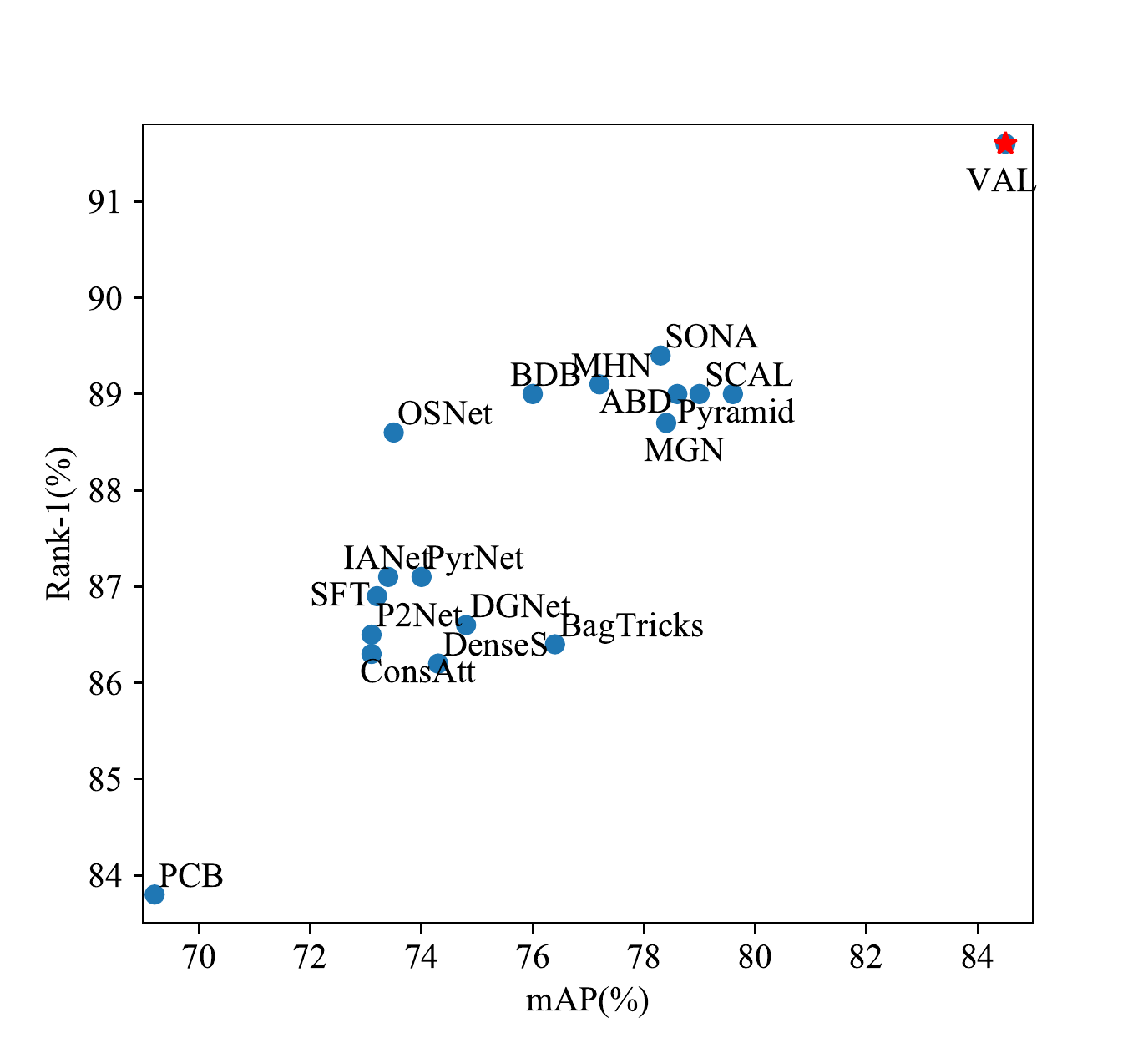}
        \centerline{\scriptsize{(b) SOTA on DukeMTMC \cite{iccv17duke} }}\medskip
        \vspace{-2mm}
    \end{minipage}%
    \begin{minipage}[t]{0.25\textwidth}
        \centering
        \vspace{-0.2cm}
        \includegraphics[height = 4.3cm,width=4.5cm]{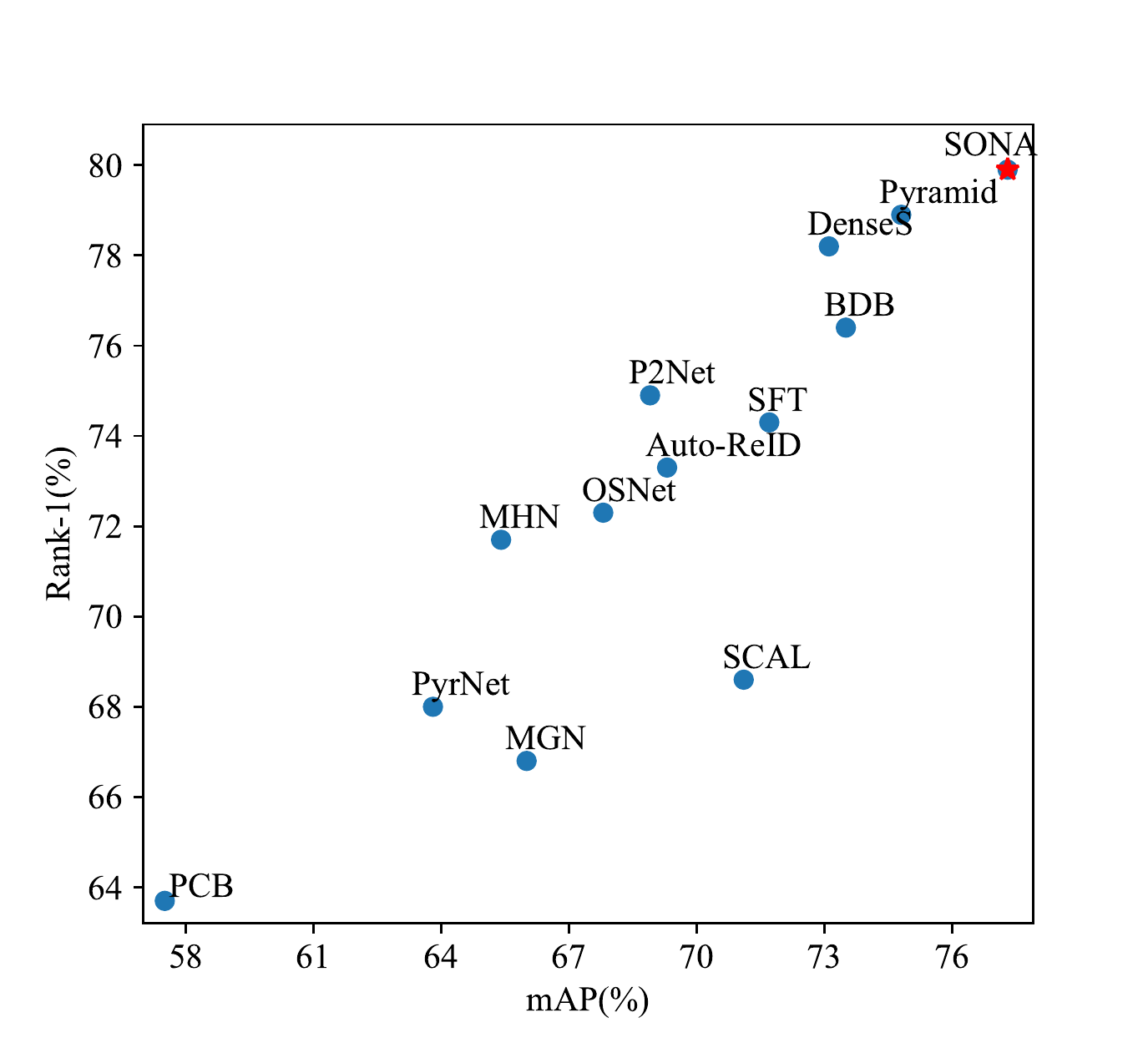}
        \vspace{-0.2cm}
        \centerline{\scriptsize{(c) SOTA on CUHK03 \cite{cvpr14cuhk} }}\medskip
        \vspace{-2mm}
    \end{minipage}%
    \begin{minipage}[t]{0.25\textwidth}
        \centering
        \vspace{-0.2cm}
        \includegraphics[height = 4.3cm,width=4.5cm]{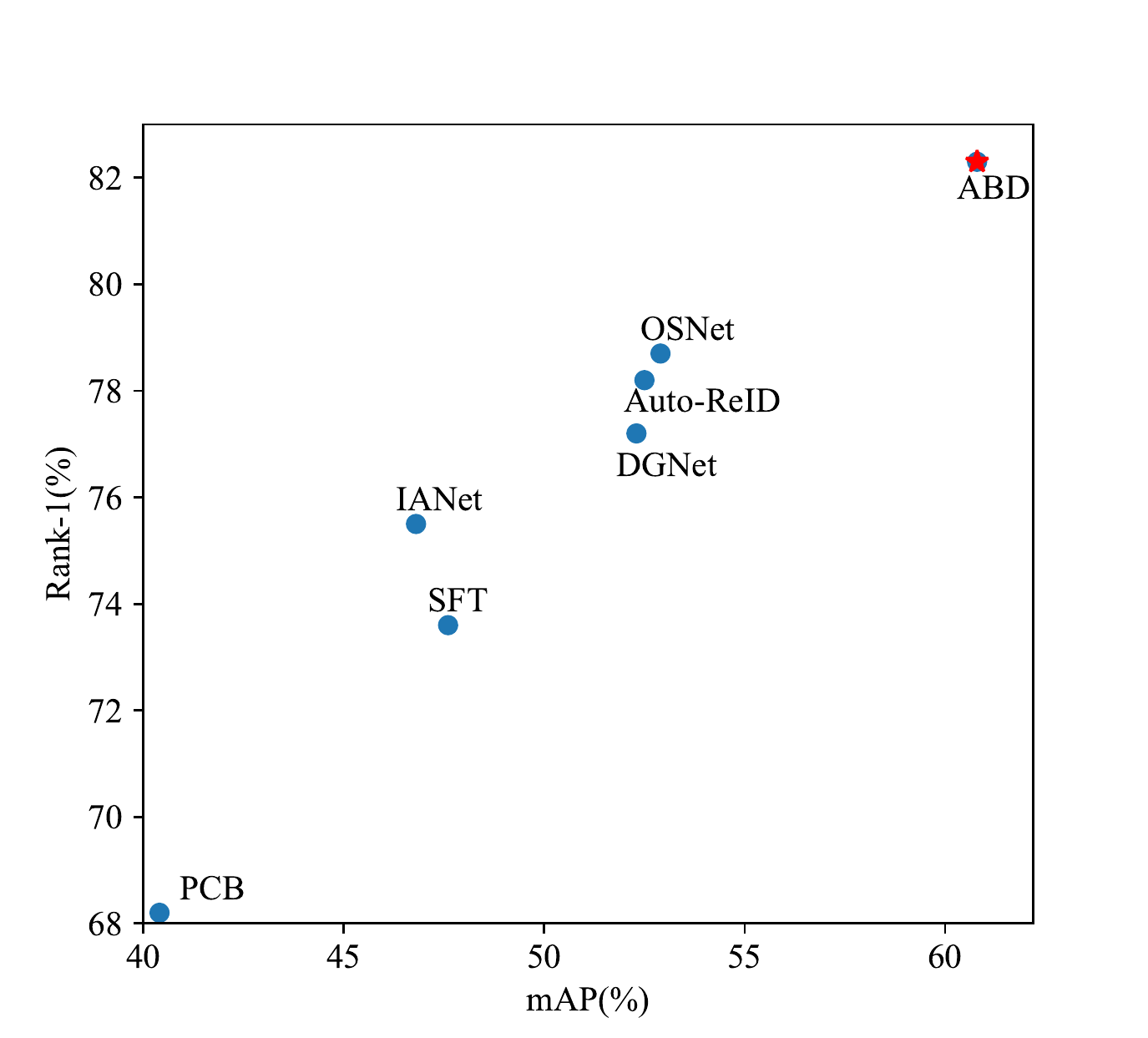}
        \centerline{\scriptsize{(d) SOTA on MSMT17 \cite{cvpr18msmt} }}\medskip
        \vspace{-2mm}
    \end{minipage}
    \vspace{-2mm}
\caption{\small{State-of-the-arts (SOTA) on four image-based person Re-ID datasets. Both the Rank-1 accuracy (\%) and mAP value (\%) are reported. For CUHK03 \cite{cvpr14cuhk}, the detected data under the setting \cite{cvpr17rerank} is reported. For Market-1501, the single query setting is used. The best result is highlighted with a {\color{red}red} star. All the listed results do not use re-ranking or additional annotated information.}}\label{fig:imgsota}
    \vspace{-2mm}
\end{figure*}

\subsection{Datasets and Evaluation}\label{sec:data}
\subsubsection{Datasets and Evaluation Metrics}\label{sec:dataset}
\textbf{Datasets.}
We first review the widely used datasets for the closed-world setting, including 11 image datasets (VIPeR \cite{eccv08elf}, iLIDS \cite{bmvc09ilids}, GRID \cite{icip13grid}, PRID2011 \cite{prid2011}, CUHK01-03 \cite{cvpr14cuhk}, Market-1501 \cite{iccv15zheng}, DukeMTMC \cite{iccv17duke}, Airport \cite{pami18survey} and MSMT17 \cite{cvpr18msmt}) and 7 video datasets (PRID-2011 \cite{prid2011}, iLIDS-VID \cite{eccv14video}, MARS \cite{eccv16mars}, Duke-Video \cite{cvpr18wu}, Duke-Tracklet \cite{eccv18untracklet}, LPW \cite{aaai18lpw} and LS-VID \cite{iccv19longshort}). The statistics of these datasets are shown in Table \ref{tab:dataset}. This survey only focuses on the general large-scale datsets for deep learning methods. A comprehensive summarization of the Re-ID datasets can be found in \cite{pami18survey} and their website\footnote{\url{https://github.com/NEU-Gou/awesome-reid-dataset}}.
Several observations can be made in terms of the dataset collection over recent years:

1) The dataset scale (both \#image and \#ID) has increased rapidly. Generally, the deep learning approach can benefit from more training samples. This also increases the annotation difficulty needed in closed-world person Re-ID. 2) The camera number is also greatly increased to approximate the large-scale camera network in practical scenarios. This also introduces additional challenges for model generalizability in a  dynamically updated network. 3) The bounding boxes generation is usually performed automatically detected/tracked, rather than mannually cropped. This simulates the real-world scenario with tracking/detection errors.

\textbf{Evaluation Metrics.}
To evaluate a Re-ID system, Cumulative Matching Characteristics (CMC) \cite{iccv07cmc} and mean Average Precision (mAP) \cite{iccv15zheng} are two widely used measurements.

CMC-$k$ (\textit{a.k.a}, Rank-$k$ matching accuracy) \cite{iccv07cmc} represents the probability that a correct match appears in the top-$k$ ranked retrieved results. CMC is accurate when only one ground truth exists for each query, since it only considers the first match in evaluation process. However, the gallery set usually contains multiple groundtruths in a large camera network, and CMC cannot completely reflect the discriminability of a model across multiple cameras.

\black{Another metric, \ie, mean Average Precision (mAP) \cite{iccv15zheng}, measures the average retrieval performance with multiple grountruths. It is originally widely used in image retrieval. For Re-ID evaluation, it can address the issue of two systems performing equally well in searching the first ground truth (might be easy match as in Fig.~\ref{fig:rerank}), but having different retrieval abilities for other hard matches.}

\black{Considering the efficiency and complexity of training a Re-ID model, some recent works \cite{iccv19osnet,iccv19autoreid} also report the FLoating-point Operations Per second (FLOPs) and the network parameter size as the evaluation metrics. These two metrics are crucial when the training/testing device has limited computational resources.}

\subsubsection{In-depth Analysis on State-of-The-Arts}\label{sec:datasota}
We review the state-of-the-arts from both image-based and video-based perspectives. We include methods published in top CV venues over the past three years.

\black{\textbf{Image-based Re-ID.} There are a large number of published papers for image-based Re-ID\footnote{\url{https://paperswithcode.com/task/person-re-identification}}. We mainly review the works published in 2019 as well as some representative works in 2018. Specifically, we include PCB \cite{eccv18pcb}, MGN \cite{mm18mgn}, PyrNet \cite{cvprw19aggregate}, Auto-ReID \cite{iccv19autoreid}, ABD-Net \cite{iccv19abdnet}, BagTricks \cite{arxiv19trick}, OSNet \cite{iccv19osnet}, DGNet \cite{cvpr19joint}, SCAL \cite{iccv19selfattention}, MHN \cite{iccv19mixattention}, P2Net \cite{iccv19dualpart}, BDB \cite{iccv19dropblock}, SONA \cite{iccv19secondnonlocal}, SFT \cite{iccv19sft}, ConsAtt \cite{iccv19consistatt}, DenseS \cite{cvpr19denselyalign}, Pyramid \cite{cvpr19multi}, IANet \cite{cvpr19interact}, VAL \cite{aaai20val}. We summarize the results on four datasets (Fig.~\ref{fig:imgsota}). This overview motivates five major insights, as discussed below.}

\black{First, with the advancement of deep learning, most of the image-based Re-ID methods have achieved higher rank-1 accuracy than humans (93.5\% \cite{arxiv17align}) on the widely used Market-1501 dataset. In particular, VAL \cite{aaai20val} obtains the best mAP of 91.6\% and Rank-1 accuracy of 96.2\% on Market-1501 dataset. The major advantage of VAL is the usage of viewpoint information. The performance can be further improved when using re-ranking or metric fusion. The success of deep learning on these closed-world datasets also motivates the shift focus to more challenging scenarios, \ie, large data size \cite{iccv19longshort} or unsupervised learning \cite{iccv17dgm}.}

Second, part-level feature learning is beneficial for discriminative Re-ID model learning. Global feature learning directly learns the representation on the whole image without the part constraints \cite{arxiv19trick}. It is discriminative when the person detection/ tracking can accurately locate the human body. When the person images suffer from large background clutter or heavy occlusions, part-level feature learning usually achieves better performance by mining discriminative body regions \cite{cvpr19partial}. Due to its advantage in handling misalignment/occlusions, we observe that most of the state-of-the-art methods developed recently adopt the features aggregation paradigm, combining the part-level and full human body features \cite{iccv19autoreid,cvpr19multi}.

Third, attention is beneficial for discriminative Re-ID model learning. We observe that all the methods (ConsAtt \cite{iccv19consistatt}, SCAL \cite{iccv19selfattention}, SONA \cite{iccv19secondnonlocal}, ABD-Net \cite{iccv19abdnet}) achieving the best performance on each dataset adopt an attention scheme. The attention captures the relationship between different convolutional channels, multiple feature maps, hierarchical layers, different body parts/regions, and even multiple images. Meanwhile,  discriminative \cite{iccv19abdnet}, diverse \cite{cvpr18diversityatt}, consistent \cite{iccv19consistatt} and high-order \cite{iccv19secondnonlocal} properties are incorporated to enhance the attentive feature learning. Considering the powerful attention schemes and the specificity of the Re-ID problem, it is highly possible that attentive deeply learned systems will continue dominating the Re-ID community, with more domain specific properties.

\black{Fourth, multi-loss training can improve the Re-ID model learning. Different loss functions optimize the network from a multi-view perspective. Combining multiple loss functions can improve the performance, evidenced by the multi-loss training strategy in the state-of-the-art methods, including ConsAtt \cite{iccv19consistatt}, ABD-Net \cite{iccv19abdnet} and SONA \cite{iccv19secondnonlocal}. In addition, a dynamic multi-loss training strategy is designed in \cite{cvpr19multi} to adaptively integrated two loss functions. The combination of identity loss and triplet loss with hard mining is the primary choice.
Moreover, due to the imbalanced issue, sample weighting strategy generally improves the performance by mining informative triplets \cite{cvpr18cascadedconv,cvpr18multicam}.}

\black{Finally, there is still much room for further improvement due to the increasing size of datasets, complex environment, limited training samples. For example, the Rank-1 accuracy (82.3\%) and mAP (60.8\%) on the newly released MSMT17 dataset \cite{cvpr18msmt} are much lower than that on Market-1501 (Rank-1: 96.2\% and mAP 91.7\%) and DukeMTMC (Rank-1: 91.6\% and mAP 84.5\%). On some other challenging datasets with limited training samples (\eg, GRID \cite{icip13grid} and VIPeR \cite{eccv08elf}), the performance is still very low. In addition, Re-ID models usually suffers significantly on cross-dataset evaluation \cite{cvpr16dgd,cvpr19general}, and the performance drops dramatically under adversarial attack \cite{iccv19attack}. We are optimistic that there would be important breakthroughs in person Re-ID, with increasing discriminability, robustness, and generalizability.}

\begin{figure*}[t]
  \centering
    \begin{minipage}[t]{0.25\textwidth}
        \centering
        \vspace{-0.2cm}
        \includegraphics[height = 4.3cm,width=4.5cm]{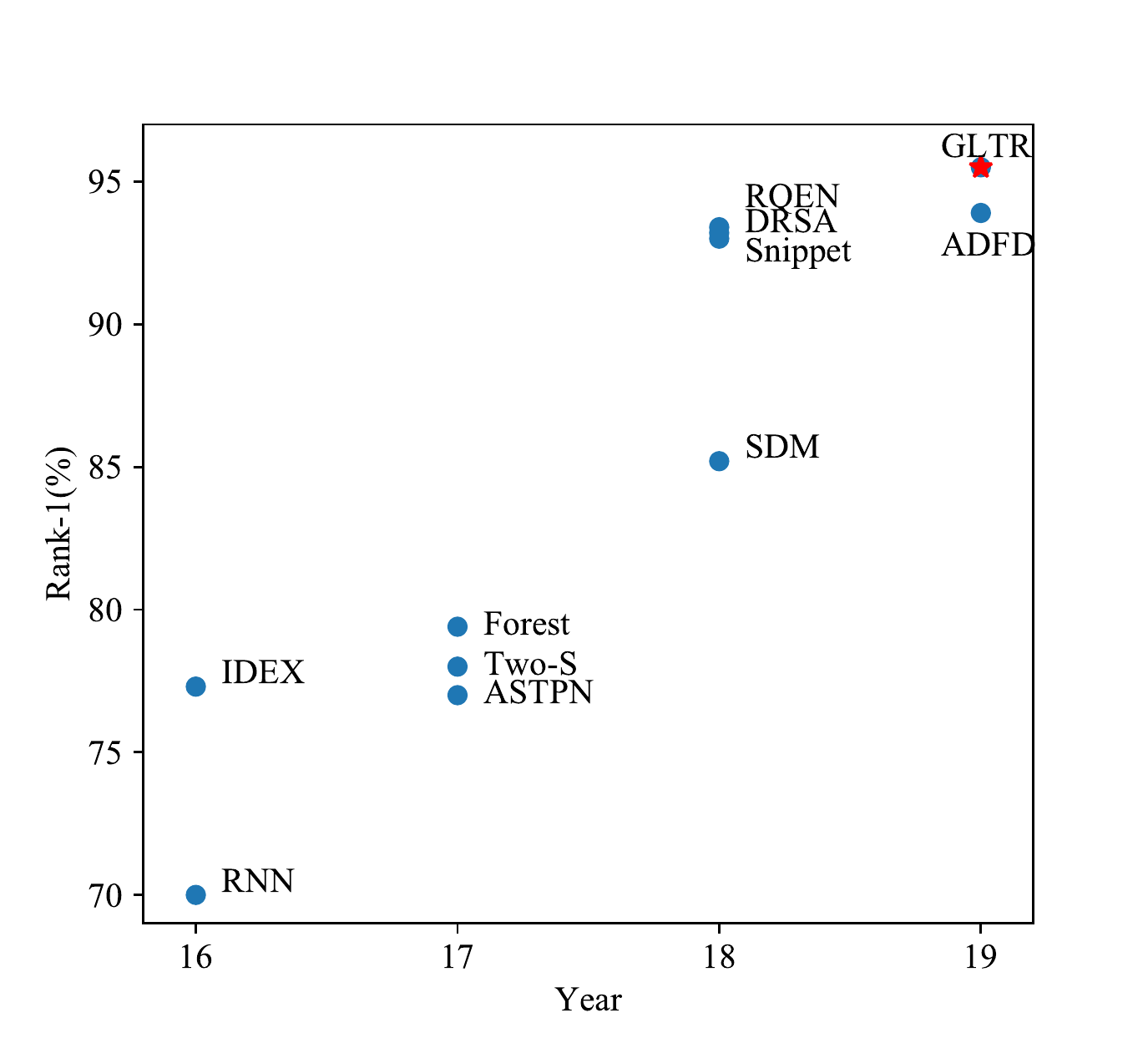}
        \vspace{-0.2cm}
        \centerline{\scriptsize{(a) SOTA on PRID-2011 \cite{prid2011}}}\medskip
        \vspace{-2mm}
    \end{minipage}%
    \begin{minipage}[t]{0.25\textwidth}
        \centering
        \vspace{-0.2cm}
        \includegraphics[height = 4.3cm,width=4.5cm]{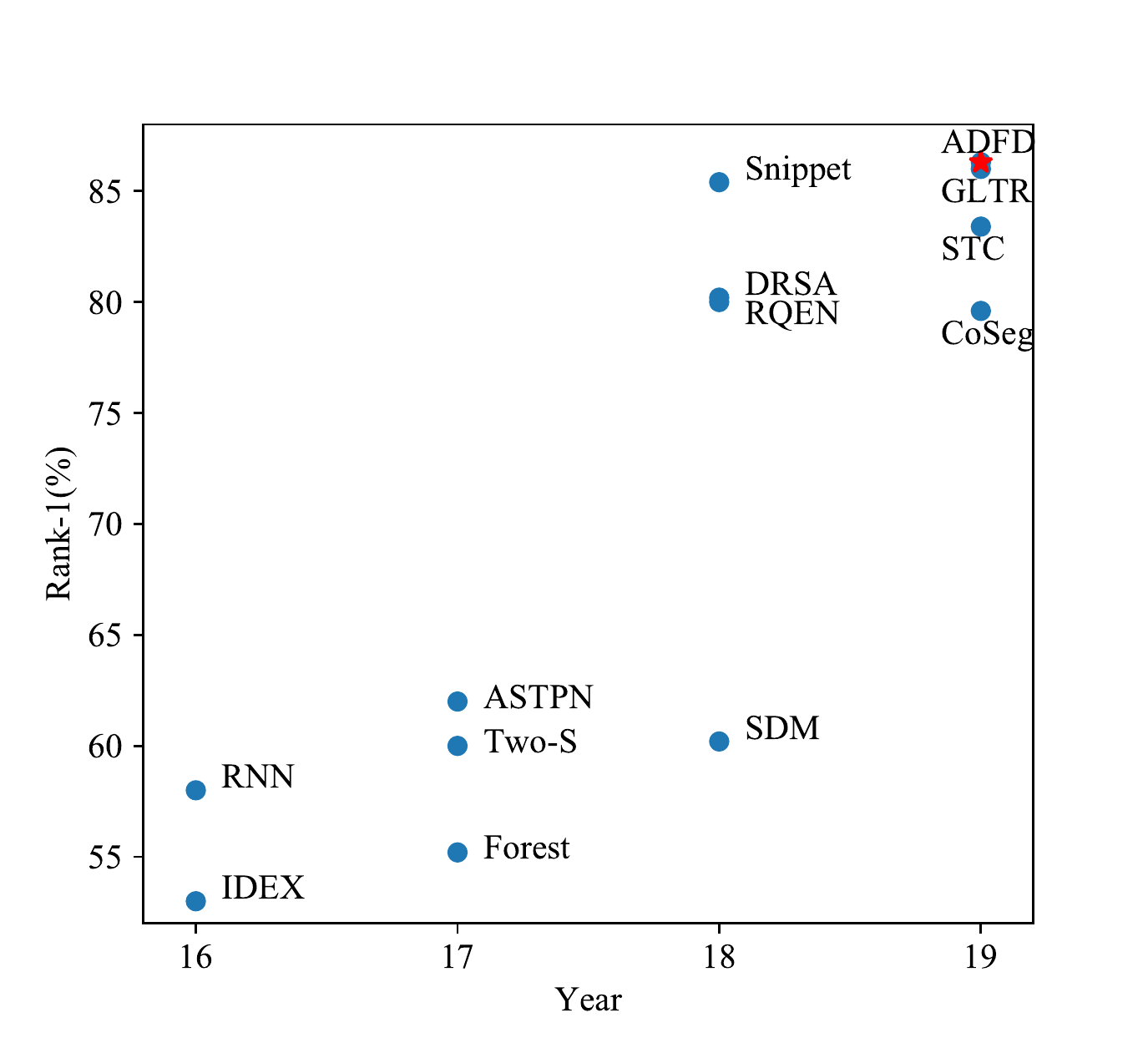}
        \centerline{\scriptsize{(b) SOTA on iLIDS-VID \cite{eccv14video}}}\medskip
        \vspace{-2mm}
    \end{minipage}%
    \begin{minipage}[t]{0.25\textwidth}
        \centering
        \vspace{-0.2cm}
        \includegraphics[height = 4.3cm,width=4.5cm]{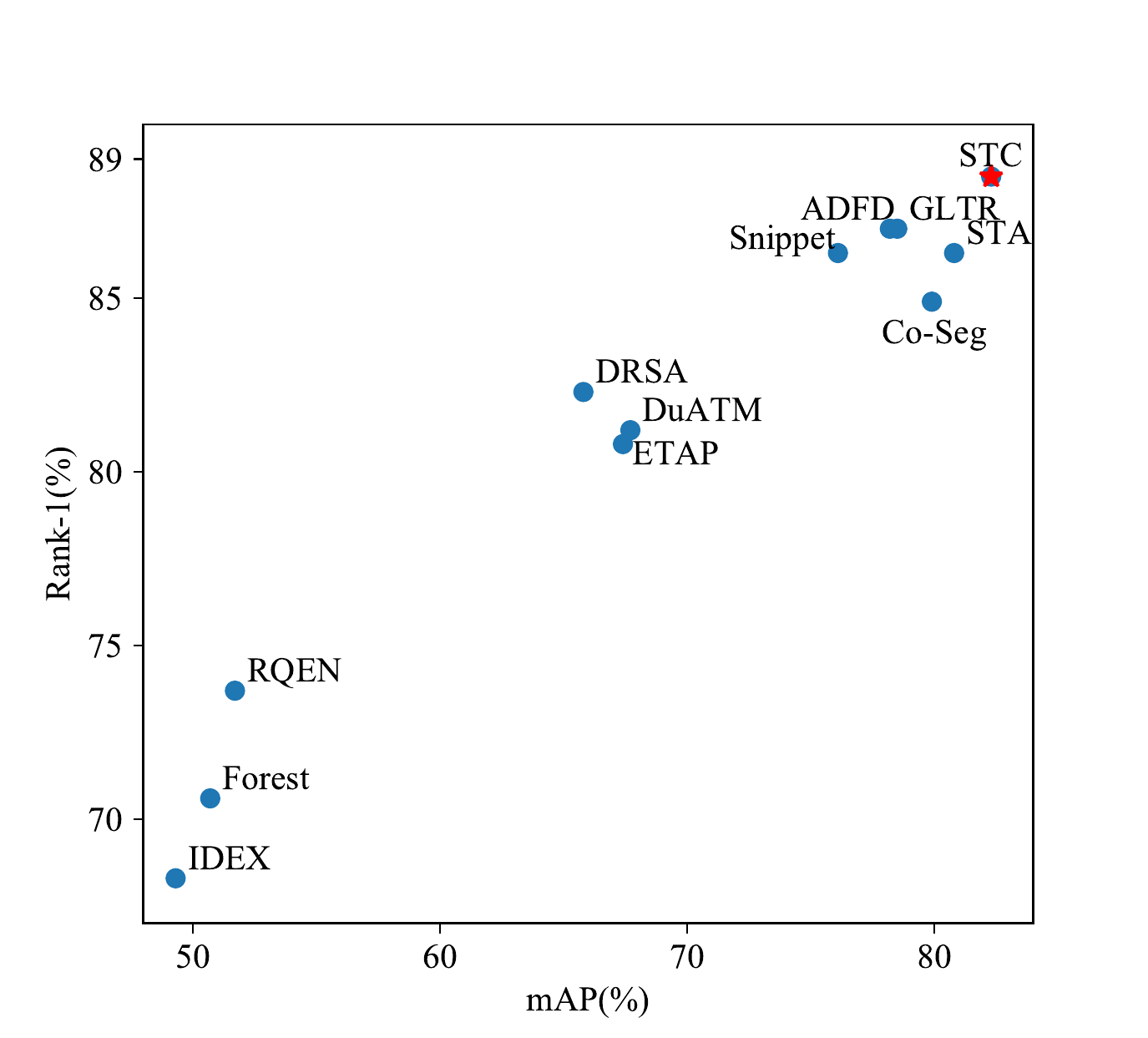}
        \vspace{-0.2cm}
        \centerline{\scriptsize{(c) SOTA on MARS \cite{eccv16mars}}}\medskip
        \vspace{-2mm}
    \end{minipage}%
    \begin{minipage}[t]{0.25\textwidth}
        \centering
        \vspace{-0.2cm}
        \includegraphics[height = 4.3cm,width=4.5cm]{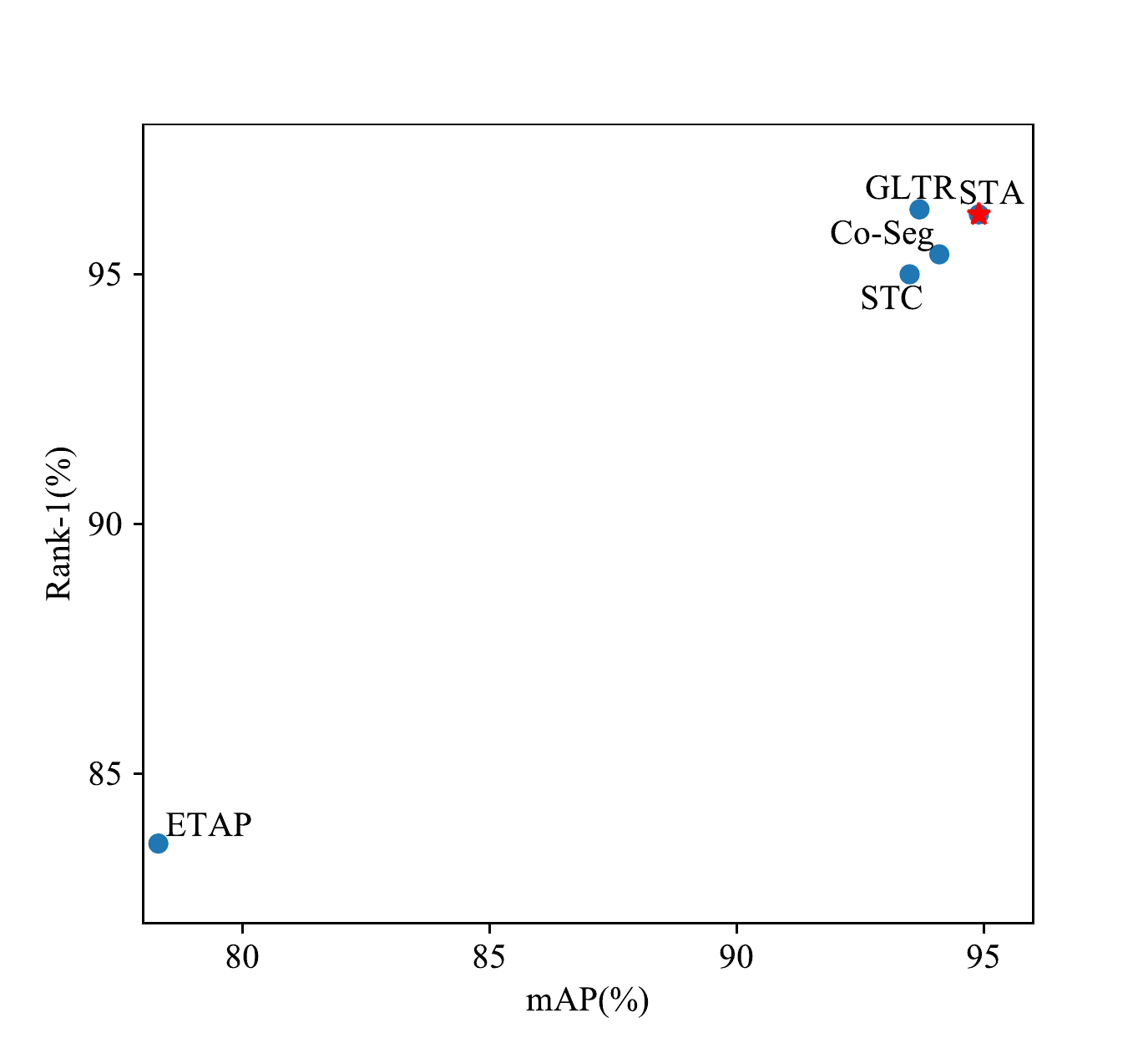}
        \centerline{\scriptsize{(d) SOTA on Duke-Video \cite{cvpr18wu}}}\medskip
        \vspace{-2mm}
    \end{minipage}
     \vspace{-2mm}
\caption{\small{State-of-the-arts (SOTA) on four widely used video-based person Re-ID datasets. The Rank-1 accuracies (\%) over years are reported. mAP values (\%) on MARS \cite{eccv16mars} and Duke-Video \cite{cvpr18wu} are reported. For Duke-Video, we refer to the settings in \cite{cvpr18wu}. The best result is highlighted with a {\color{red}red} star. All the listed results do not use re-ranking or additional annotated information.}}\label{fig:sotavid}
 \vspace{-2mm}
\end{figure*}

 \textbf{Video-based Re-ID.}
Video-based Re-ID has received less interest, compared to image-based Re-ID. We review the deeply learned Re-ID models, including CoSeg \cite{iccv19coseg}, GLTR \cite{iccv19longshort}, STA \cite{aaai19sta}, ADFD \cite{cvpr19videoatt}, STC \cite{cvpr19occlusion}, DRSA \cite{cvpr18diversityatt}, Snippet \cite{cvpr18snippet}, ETAP
\cite{cvpr18wu}, DuATM \cite{cvpr18dualatt}, SDM \cite{cvpr18decisionmaking}, TwoS \cite{iccv17videotwo}, ASTPN \cite{iccv17joint}, RQEN \cite{aaai18lpw}, Forest \cite{cvpr17forest}, RNN \cite{cvpr16video} and IDEX \cite{eccv16mars}. We also summarize the results on four video Re-ID datasets, as shown in Fig.~\ref{fig:sotavid}. From these results, the following observations can be drawn.

\black{First, a clear trend of increasing performance can be seen over the years with the development of deep learning techniques. Specifically, the Rank-1 accuracy increases from 70\% (RNN \cite{cvpr16video} in 2016) to 95.5\% (GLTR \cite{iccv19longshort} in 2019) on PRID-2011 dataset, and from 58\% (RNN \cite{cvpr16video}) to 86.3\% (ADFD \cite{cvpr19videoatt}) on iLIDS-VID dataset. On the large-scale MARS dataset, the Rank-1 accuracy/mAP increase from 68.3\%/49.3\% (IDEX \cite{eccv16mars}) to 88.5\%/82.3\% (STC \cite{cvpr19occlusion}). On the Duke-Video dataset \cite{cvpr18wu}, STA \cite{aaai19sta} also achieves a Rank-1 accuracy of 96.2\%, and the mAP is 94.9\%.}

Second, spatial and temporal modeling is crucial for discriminative video representation learning. We observe that all the methods (STA \cite{aaai19sta}, STC \cite{cvpr19occlusion}, GLTR \cite{iccv19longshort}) design spatial-temporal aggregation strategies to improve the video Re-ID performance. Similar to image-based Re-ID, the attention scheme across multiple frames \cite{aaai19sta,cvpr19videoatt} also greatly enhances the discriminability. Another interesting observation in \cite{cvpr19occlusion} demonstrates that utilizing multiple frames within the video sequence can fill in the occluded regions, which provides a possible solution for handling the challenging occlusion problem in the future.

Finally, the performance on these datases has reached a saturation state, usually about less than 1\% accuracy gain on these four video datasets. However, there is still large room for improvements on the challenging cases. For example, on the newly collected video dataset, LS-VID \cite{iccv19longshort}, the Rank-1 accuracy/mAP of GLTR \cite{iccv19longshort} are only 63.1\%/44.43\%, while GLTR \cite{iccv19longshort} can achieve state-of-the-art or at least comparable performance on the other four daatsets. LS-VID \cite{iccv19longshort} contains significantly more identities and video sequences. This provides a challenging benchmark for future breakthroughs in video based Re-ID.

\section{Open-world Person Re-Identification}\label{sec:open}
This section reviews open-world person Re-ID as discussed in \S~\ref{sec:intro}, including heterogeneous Re-ID by matching person images across heterogeneous modalities (\S~\ref{sec:hetro}), end-to-end Re-ID from the raw images/videos (\S~\ref{sec:end2end}), semi-/unsupervised learning with limited/unavailable annotated labels (\S~\ref{sec:unsuper}), robust Re-ID model learning with noisy annotations (\S~\ref{sec:noise}) and open-set person Re-ID when the correct match does not occur in the gallery (\S~\ref{sec:openset}).
\subsection{Heterogeneous Re-ID}\label{sec:hetro}
This subsection summarizes four main kinds of heterogeneous Re-ID, including Re-ID between depth and RGB images (\S~\ref{sec:depth}), text-to-image Re-ID (\S~\ref{sec:text2img}), visible-to-infrared Re-ID (\S~\ref{sec:vireid}) and cross resolution Re-ID (\S~\ref{sec:crossresolution}).
\subsubsection{Depth-based Re-ID}\label{sec:depth}
Depth images capture the body shape and skeleton information. This provides the possibility for Re-ID under illumination/clothes changing environments, which is also important for personalized human interaction applications.

A recurrent attention-based model is proposed in \cite{cvpr16depth} to address the depth-based person identification. In a reinforcement learning framework, they combine
the convolutional and recurrent neural networks to identify small, discriminative local regions of the human body.

Karianakis \textit{et al.} \cite{eccv18depth} leverage the large RGB datasets to design a split-rate RGB-to-Depth transfer method, which bridges the gap between the depth images and the RGB images. Their model further incorporates a temporal attention to enhance video representation for depth Re-ID.

Some methods \cite{tip17rgbd,eccv12rgbd} have also studied the combination of RGB and depth information to improve the Re-ID performance, addressing the clothes-changing challenge.

\subsubsection{Text-to-Image Re-ID}\label{sec:text2img}
Text-to-image Re-ID addresses the matching between a text description and RGB images \cite{cvpr17nlp}. It is imperative when the visual image of query person cannot be obtained, and only a text description can be alternatively provided.

\black{A gated neural attention model \cite{cvpr17nlp} with recurrent neural network learns the shared features between the text description and the person images. This enables the end-to-end training for text to image pedestrian retrieval.
Cheng \textit{et al.} \cite{eccv18imgtext} propose a global discriminative image-language association learning method, capturing the identity discriminative information and local reconstructive image-language association under a reconstruction process. A cross projection learning method \cite{eccv18img2text} also learns a shared space with image-to-text matching. A deep adversarial graph attention convolution network is designed in \cite{mm19img2txt} with graph relation mining. However, the large semantic gap between the text descriptions and the visual images is still challenging. Meanwhile, how to combine the texts and hand-painting sketch image is also worth studying in the future.}


\subsubsection{Visible-Infrared Re-ID}\label{sec:vireid}
Visible-Infrared Re-ID handles the cross-modality matching between the daytime visible and night-time infrared images. It is important in low-lighting conditions, where the images can only be captured by infrared cameras \cite{iccv17cross,sensors17,ijcai18vtreid}.

\blue{Wu \textit{et al.} \cite{iccv17cross} start the first attempt to address this issue, by proposing a deep zero-padding framework \cite{iccv17cross} to adaptively learn the modality sharable features. A two stream network is introduced in \cite{aaai18vtreid,tifs19vtreid} to model the modality-sharable and -specific information, addressing the intra- and cross-modality variations simultaneously. Besides the cross-modality shared embedding learning \cite{aaai19hsme}, the classifier-level discrepancy is also investigated in \cite{tifs20gray}. Recent methods \cite{cvpr19ivreid,iccv19ivreid} adopt the GAN technique to generate cross-modality person images to reduce the cross-modality discrepancy at both image and feature level. Hierarchical cross-Modality disentanglement factors are modeled in \cite{cvpr20hicmd}. A dual-attentive aggregation learning method is presented in \cite{eccv20ddag} to capture multi-level relations.}


\subsubsection{Cross-Resolution Re-ID}\label{sec:crossresolution}
Cross-Resolution Re-ID conducts the matching between low-resolution and high-resolution images, addressing the large resolution variations \cite{iccv15lowres,cvpr18multires}.
A cascaded SR-GAN \cite{ijcai18resolution} generates the high-resolution person images in a cascaded manner, incorporating the identity information. Li \textit{et al.} \cite{iccv19crossres} adopt the adversarial learning technique to obtain resolution-invariant image representations.

\subsection{End-to-End Re-ID}\label{sec:end2end}
End-to-end Re-ID alleviates the reliance on additional step for bounding boxes generation. It involves the person Re-ID from raw images or videos, and multi-camera tracking.

\textbf{Re-ID in Raw Images/Videos}
This task requires that the model jointly performs the person detection and re-identification in a single framework \cite{cvpr17joint,cvpr17prw}. It is challenging due to the different focuses of two major components.

\blue{Zheng \textit{et al.} \cite{cvpr17prw} present a two-stage framework, and systematically evaluate the benefits and limitations of person detection for the later stage person Re-ID.
Xiao \textit{et al.} \cite{cvpr17joint} design an end-to-end person search system using a single convolutional neural network for joint person detection and re-identification.
A Neural Person Search Machine (NPSM) \cite{iccv17neuralmachine} is developed to recursively refine the searching area and locate the target person by fully exploiting the contextual information between the query and the detected candidate region. Similarly, a contextual instance expansion module \cite{cvpr19context} is learned in a graph learning framework to improve the end-to-end person search. A query-guided end-to-end person search system \cite{cvpr19queryguide} is developed using the Siamese squeeze-and-excitation network to capture the global context information with query-guided region proposal generation. A localization refinement scheme with discriminative Re-ID feature learning is introduced in \cite{iccv19locarefine} to generate more reliable bounding boxes. An Identity DiscriminativE Attention reinforcement Learning (IDEAL) method \cite{bmvc17attention} selects informative regions for auto-generated bounding boxes, improving the Re-ID performance.}

Yamaguchi \textit{et al.} \cite{iccv17endvideo} investigate a more challenging problem, \ie, searching for the person from raw videos with text description. A multi-stage method with spatio-temporal person detection and multi-modal retrieval is proposed. Further exploration along this direction is expected.

\textbf{Multi-camera Tracking}
End-to-end person Re-ID is also closely related to multi-person, multi-camera tracking \cite{cvpr18multicam}. A graph-based formulation to link person hypotheses is proposed for multi-person tracking \cite{cvpr17multiple}, where the holistic features of the full human body and body pose layout are combined as the representation for each person.
Ristani \textit{et al.} \cite{cvpr18multicam} learn the correlation between the multi-target multi-camera tracking and person Re-ID by hard-identity mining and adaptive weighted triplet learning. Recently, a locality aware appearance metric (LAAM) \cite{arxiv19multicam} with both intra- and inter-camera relation modeling is proposed.

\subsection{Semi-supervised and Unsupervised Re-ID}\label{sec:unsuper}

\subsubsection{Unsupervised Re-ID}\label{sec:unsuperun}
\black{Early unsupervised Re-ID mainly learns invariant components, \ie, dictionary \cite{eccv16un}, metric \cite{iccv17smp} or saliency \cite{cvpr13saliency}, which leads to limited discriminability or scalability.}

\black{For deeply unsupervised methods, cross-camera label estimation is one the popular approaches \cite{iccv17dgm,arxiv17uncluster}. Dynamic graph matching (DGM) \cite{tip19dgm} formulates the label estimation as a bipartite graph matching problem. To further improve the performance, global camera network
constraints \cite{arxiv19camera} are exploited for consistent matching. Liu \textit{et al.} progressively mine the labels with step-wise metric promotion \cite{iccv17smp}. A robust anchor embedding method \cite{eccv18race} iteratively assigns labels to the unlabelled tracklets to enlarge the anchor video sequences set. With the estimated labels, deep learning can be applied to learn Re-ID models.}

\blue{For end-to-end unsupervised Re-ID, an iterative clustering and Re-ID model learning is presented in \cite{arxiv17uncluster}. Similarly, the relations among samples are utilized in a
hierarchical clustering framework \cite{cvpr20hct}. Soft multi-label learning \cite{cvpr19unsoft} mines the soft label information from a reference set for unsupervised learning. A Tracklet Association Unsupervised Deep Learning (TAUDL) framework \cite{eccv18untracklet} jointly conducts the within-camera tracklet association and model the cross-camera tracklet correlation. Similarly, an unsupervised camera-aware similarity consistency mining method \cite{iccv19unwu} is also presented in a coarse-to-fine consistency learning scheme. The intra-camera mining and inter-camera association is applied in a graph association framework \cite{iccv19ungraph}. The semantic attributes are also adopted in Transferable Joint Attribute-Identity Deep Learning (TJ-AIDL) framework \cite{cvpr18att}. However, it is still challenging for model updating with newly arriving unlabelled data.}

\blue{Besides, several methods have also tried to learn a part-level representation based on the observation that it is easier to mine the label information in local parts than that of a whole image. A PatchNet \cite{cvpr19unpatch} is designed to learn discriminative patch features by mining patch level similarity. A Self-similarity Grouping (SSG) approach \cite{iccv19selfgroup} iteratively conducts grouping (exploits both the global body and local parts similarity for pseudo labeling) and Re-ID model training in a self-paced manner.}

\textbf{Semi-/Weakly supervised Re-ID.}
With limited label information, a one-shot metric learning method is proposed in \cite{cvpr17oneshot}, which incorporates a deep texture representation and a color metric.
A stepwise one-shot learning method (EUG) is proposed in \cite{cvpr18wu} for video-based Re-ID, gradually selecting a few candidates from unlabeled tracklets to enrich the labeled tracklet set. A multiple instance attention learning framework \cite{arxiv20weak} uses the video-level labels for representation learning, alleviating the reliance on full annotation.

\subsubsection{Unsupervised Domain Adaptation}\label{sec:unsuperuda}
Unsupervised domain adaptation (UDA) transfers the knowledge on a labeled source dataset to the unlabeled target dataset \cite{iccv13domain}. Due to the large domain shift and powerful supervision in source dataset, it is another popular approach for unsupervised Re-ID without target dataset labels.

\blue{\textbf{Target Image Generation.} Using GAN generation to transfer the source domain images to target-domain style is a popular approach for UDA Re-ID. With the generated images, this enables supervised Re-ID model learning in the unlabeled target domain. Wei \textit{et al.} \cite{cvpr18msmt} propose a Person Transfer Generative Adversarial Network (PTGAN), transferring the knowledge from one labeled source dataset to the unlabeled target dataset. Preserved self-similarity and domain-dissimilarity \cite{cvpr18spgan} is trained with a similarity preserving generative adversarial network (SPGAN). A Hetero-Homogeneous Learning (HHL) method \cite{eccv18zhong} simultaneously considers the camera invariance with homogeneous learning and domain connectedness with heterogeneous learning. An adaptive transfer network \cite{cvpr19adaptrans} decomposes the adaptation process into certain imaging factors, including illumination, resolution, camera view, etc. This strategy improves the cross-dataset performance. Huang \textit{et al.} \cite{iccv19sbsgan} try to suppress the background shift to minimize the domain shift problem. Chen \textit{et al.} \cite{iccv19inscontext} design an instance-guided context rendering scheme to transfer the person identities from source domain into diverse contexts in the target domain. Besides, a pose disentanglement scheme is added to improve the image generation \cite{iccv19crosspose}.  A mutual mean-teacher learning scheme is also developed in \cite{iclr20mmt}. However, the scalability and stability of the image generation for practical large-scale changing environment are still challenging.}

\blue{Bak \textit{et al.} \cite{eccv18bak} generate a synthetic dataset with different illumination conditions to model realistic indoor and outdoor lighting. The synthesized dataset increases generalizability of the learned model and can be easily adapted to a new dataset without additional supervision \cite{mm20synth}.}

\blue{\textbf{Target Domain Supervision Mining.} Some methods directly mine the supervision on the unlabeled target dataset with a well trained model from source dataset. An exemplar memory learning scheme \cite{cvpr19exemplarmemory} considers three invariant cues as the supervision, including exemplar-invariance, camera invariance and neighborhood-invariance. The Domain-Invariant Mapping Network (DIMN) \cite{cvpr19general} formulates a meta-learning pipeline for the domain transfer task, and a subset of source domain is sampled at each training episode to update the memory bank, enhancing the scalability and discriminability.
The camera view information is also applied in \cite{iccv19uncamera} as the supervision signal to reduce the domain gap.
A self-training method with progressive augmentation \cite{iccv19progressiveaug} jointly captures the local structure and global data distribution on the target dataset. Recently, a self-paced contrastive learning framework with hybrid memory \cite{nips20spcl} is developed with great success, which dynamically generates multi-level supervision signals.}

\blue{The spatio-temporal information is also utilized as the supervision in TFusion \cite{cvpr18tffusion}. TFusion transfers the spatio-temporal patterns learned in the source domain to the target domain with a Bayesian fusion model. Similarly, Query-Adaptive Convolution (QAConv) \cite{eccv20liao} is developed to improve cross-dataset accuracy.}

\setlength{\tabcolsep}{4.5pt}
\begin{table}[t]\scriptsize
\centering
\caption{Statistics of SOTA unsupervised person Re-ID on two image-based datasets. ``Source" represents if it utilizes the source annotated data in training the target Re-ID model. ``Gen." indicates if it contains an image generation process. Rank-1 accuracy (\%) and mAP (\%) are reported.}
\vspace{-2mm}
 \begin{threeparttable}
\begin{tabular}{l|P{0.8cm}P{0.6cm}|P{0.6cm}P{0.6cm}|P{0.6cm}P{0.6cm}}
\hline
   & && \multicolumn{2}{c}{\textit{Market-1501}} & \multicolumn{2}{c}{\textit{DukeMTMC}}\\
  Methods                  & Source & Gen.  & R1    & mAP  & R1  & mAP \\\hline
  CAMEL~\cite{iccv17unmetric} \tiny{ICCV17} & Model & No & 54.5 &26.3 &  -&- \\
 PUL~\cite{arxiv17uncluster} \tiny{TOMM18} & Model & No & 45.5 & 20.5 & 30.0 & 16.4 \\
 PTGAN~\cite{cvpr18spgan} \tiny{CVPR18}  & Data &  Yes &58.1   & 26.9  & 46.9  & 26.4\\
 TJ-AIDL\tnote{\dag}~\cite{cvpr18att} \tiny{CVPR18}  & Data & No & 58.2 &26.5 & 44.3 & 23.0 \\
 HHL~\cite{eccv18zhong} \tiny{ECCV18} & Data & Yes & 62.2 &31.4 & 46.9 & 27.2 \\
 MAR\tnote{\ddag} \ \cite{cvpr19unsoft}  \tiny{CVPR19}  & Data & No & 67.7 & 40.0 & 67.1 & 48.0 \\
 ENC~\cite{cvpr19exemplarmemory} \tiny{CVPR19}  & Data & No & 75.1  & 43.0 & 63.3 & 40.4 \\
 ATNet~\cite{cvpr19adaptrans} \tiny{CVPR19}  & Data & Yes & 55.7 & 25.6 & 45.1 & 24.9 \\
 PAUL\tnote{\ddag} \ \cite{cvpr19unpatch} \tiny{CVPR19}  & Model & No & 68.5 & 40.1 & 72.0 & 53.2 \\
 SBGAN~\cite{iccv19sbsgan}  \tiny{ICCV19} & Data & Yes & 58.5 & 27.3 & 53.5 & 30.8 \\
 UCDA~\cite{iccv19uncamera} \tiny{ICCV19} & Data & No & 64.3 & 34.5 & 55.4 & 36.7 \\
 CASC\tnote{\ddag} \ \cite{iccv19unwu} \tiny{ICCV19} & Model & No & 65.4 & 35.5  & 59.3 & 37.8 \\
  PDA~\cite{iccv19crosspose} \tiny{ICCV19} & Data & Yes & 75.2 & 47.6 & 63.2 & 45.1 \\
 CR-GAN~\cite{iccv19inscontext} \tiny{ICCV19} & Data & Yes & 77.7 &  54.0 & 68.9 & 48.6 \\
  PAST~\cite{iccv19progressiveaug} \tiny{ICCV19} & Model & No & 78.4 &54.6 &72.4 & 54.3 \\
 SSG~\cite{iccv19selfgroup} \tiny{ICCV19}   & Model & No & 80.0 & 58.3 & 73.0 & 53.4 \\
 HCT~\cite{cvpr20hct} \tiny{CVPR20}& Model & No & 80.0 & 56.4 & 69.6 & 50.7 \\
 SNR~\cite{cvpr20snr} \tiny{CVPR20} & Data & No & 82.8 & 61.7 & 76.3 & 58.1 \\
 MMT~\cite{iclr20mmt} \tiny{ICLR20} & Data & No & 87.7 & 71.2 & 78.0 & 65.1 \\
 MEB-Net~\cite{eccv20meb} \tiny{ECCV20} & Data & No & 89.9 & 76.0 & 79.6 & 66.1 \\
 SpCL~\cite{nips20spcl} \tiny{NeurIPS20} & Data & No & \textbf{90.3} & \textbf{76.7} &\textbf{82.9} &\textbf{68.8} \\\hline
\end{tabular}
        \scriptsize{$\bullet$ $^\dag$ TJ-AIDL~\cite{cvpr18att} requires additional attribute annotation.}\\
        \scriptsize{$\bullet$ $^\S$ DAS~\cite{eccv18bak} generates synthesized virtual humans under vairous lightings.}\\
       \scriptsize{$\bullet$ $^\ddag$ PAUL~\cite{cvpr19unpatch}, MAR~\cite{cvpr19unsoft} and CASC~\cite{iccv19unwu} use MSMT17 as source dataset.}\\
 \end{threeparttable}
\label{tab:sotaun}
\vspace{-5mm}
\end{table}

\subsubsection{State-of-The-Arts for Unsupervised Re-ID}\label{sec:unsota}
\blue{Unsupervised Re-ID has achieved increasing attention in recent years, evidenced by the increasing number of publications in top venues. We review the SOTA for unsupervised deeply learned methods on two widely-used image-based Re-ID datasets.
The results are summarized in Table \ref{tab:sotaun}. From these results, the following insights can be drawn.}

\blue{First, the unsupervised Re-ID performance has increased significantly over the years. The Rank-1 accuracy/mAP increases from 54.5\%/26.3\% (CAMEL \cite{iccv17unmetric}) to 90.3\%/76.7\% (SpCL \cite{nips20spcl}) on the Market-1501 dataset within three years. The performance for DukeMTMC dataset increases from 30.0\%/16.4\% to 82.9\%/68.8\%. The gap between the supervised upper bound and the unsupervised learning is narrowed significantly. This demonstrates the success of unsupervised Re-ID with deep learning.}

\blue{Second, current unsupervised Re-ID is still under-developed and it can be further improved in the following aspects: 1) The powerful attention scheme in supervised Re-ID methods has rarely been applied in unsupervised Re-ID. 2) Target domain image generation has been proved effective in some methods, but they are not applied in two best methods (PAST \cite{iccv19progressiveaug}, SSG \cite{iccv19selfgroup}). 3) Using the annotated source data in the training process of the target domain is beneficial for cross-dataset learning, but it is also not included in above two methods. These observations provide the potential basis for further improvements.}

\blue{Third, there is still a large gap between the unsupervised and supervised Re-ID. For example, the rank-1 accuracy of supervised ConsAtt \cite{iccv19consistatt} has achieved 96.1\% on the Market-1501 dataset, while the highest accuracy of unsupervised SpCL \cite{nips20spcl} is about 90.3\%. Recently, He \textit{et al.} \cite{arxiv19momentum} have demonstrated that unsupervised learning with large-scale unlabeled training data has the ability to outperform the supervised learning on various tasks \cite{pami20embedding}. We expect that several breakthroughs in future unsupervised Re-ID.}

\subsection{Noise-Robust Re-ID}\label{sec:noise}
Re-ID usually suffers from unavoidable noise due to data collection and annotation difficulty. We review noise-robust Re-ID from three aspects: \textit{Partial Re-ID} with heavy occlusion, \textit{Re-ID with sample noise} caused by detection or tracking errors, and \textit{Re-ID with label noise} caused by annotation error.

\black{\textbf{Partial Re-ID.}
This addresses the Re-ID problem with heavy occlusions, \ie, only part of the human body is visible \cite{iccv15partial}.
A fully convolutional network \cite{cvpr18partial} is adopted to generate fix-sized spatial feature maps for the incomplete person images. Deep Spatial feature Reconstruction (DSR) is further incorporated to avoid explicit alignment by exploiting the reconstructing error. Sun \textit{et al.} \cite{cvpr19partial} design a Visibility-aware Part Model (VPM) to extract sharable region-level features, thus suppressing the spatial misalignment in the incomplete images. A foreground-aware pyramid reconstruction scheme \cite{iccv19foreground} also tries to learn from the unoccluded regions.
The Pose-Guided Feature Alignment (PGFA) \cite{iccv19partial} exploits the pose landmarks to mine discriminative part information from occlusion noise. However, it is still challenging due to the severe partial misalignment, unpredictable visible regions and distracting unshared body regions. Meanwhile, how to adaptively adjust the matching model for different queries still needs further investigation.}

\black{\textbf{Re-ID with Sample Noise.}
This refers to the problem of the person images or the video sequence containing outlying regions/frames, either caused by poor detection/inaccurate tracking results.
To handle the outlying regions or background clutter within the person image, pose estimation cues \cite{cvpr17spindle,cvpr18pose} or attention cues \cite{bmvc17attention,cvpr18mask,cvpr13saliency} are exploited. The basic idea is to suppress the contribution of the noisy regions in the final holistic representation. For video sequences, set-level feature learning \cite{eccv18race} or frame level re-weighting \cite{cvpr18snippet} are the commonly used approaches to reduce the impact of noisy frames. Hou \textit{et al.} \cite{cvpr19occlusion} also utilize multiple video frames to auto-complete occluded regions. It is expected that more domain-specific sample noise handling designs in the future.}


\black{\textbf{Re-ID with Label Noise.}
Label noise is usually unavoidable due to annotation error. Zheng \textit{et al.} adopt a label smoothing technique to avoid label overfiting issues \cite{iccv17duke}. A Distribution Net (DNet) that models the feature uncertainty is proposed in \cite{iccv19dnet} for robust Re-ID model learning against label noise, reducing the impact of samples with high feature uncertainty. Different from the general classification problem, robust Re-ID model learning suffers from limited training samples for each identity \cite{tifs20noisy}. In addition, the unknown new identities increase additional difficulty for the robust Re-ID model learning.}

\subsection{Open-set Re-ID and Beyond}\label{sec:openset}
Open-set Re-ID is usually formulated as a person verification problem, \ie, discriminating whether or not two person images belong to the same identity \cite{icip16openset,tip18open}. The verification usually requires a learned condition $\tau$, \ie,  $sim(query, gallery) > \tau$. Early researches design hand-crafted systems \cite{icip16openset,tip18open,pami15openworld}. For deep learning methods,
an Adversarial PersonNet (APN) is proposed in \cite{eccv18open}, which jointly learns a GAN module and the Re-ID feature extractor. The basic idea of this GAN is to generate realistic target-like images (imposters) and enforce the feature extractor is robust to the generated image attack. Modeling feature uncertainty is also investigated in \cite{iccv19dnet}. However, it remains quite challenging to achieve a high true target recognition and maintain low false target recognition rate \cite{pami97error}.

\black{\textbf{Group Re-ID.}
It aims at associating the persons in groups  rather than individuals \cite{bmvc09ilids}. Early researches mainly focus on group representation extraction with sparse dictionary learning \cite{iccv17groupreid} or covariance
descriptor aggregation \cite{icpr10group}. The multi-grain information is integrated in \cite{mm18group} to fully capture the characteristics of
a group. Recently, the graph convoltuional network is applied in \cite{mm19group}, representing the group as a graph. The group similarity is also applied in the end-to-end person search \cite{cvpr19context} and the individual re-identification \cite{eccv18graphsim,cvpr19queryguide} to improve the accuracy. However, group Re-ID is still challenging since the group variation is more complicated than the individuals.}

\black{\textbf{Dynamic Multi-Camera Network.}
Dynamic updated multi-camera network is another challenging issue \cite{icip15active,eccv16temporal,cviu17camera,eccv14camera}, which needs model adaptation for new cameras or probes. A human in-the-loop incremental learning method is introduced in \cite{eccv16temporal} to update the Re-ID model, adapting the representation for different probe galleries. Early research also applies the active learning \cite{icip15active} for continuous Re-ID in multi-camera network. A continuous adaptation method based on sparse non-redundant representative selection is introduced in \cite{cviu17camera}. A transitive inference algorithm \cite{cvpr17dynamicopen} is designed to exploit the best source camera model based on a geodesic flow kernel. Multiple environmental constraints (\eg, Camera Topology) in dense crowds and social relationships are integrated for an open-world person Re-ID system \cite{eccv16reidcrowd}. The model adaptation and environmental factors of cameras are crucial in practical dynamic multi-camera network. Moreover, how to apply the deep learning technique for the dynamic multi-camera network is still less investigated.}


\begin{figure}[t]
  \centering
  \includegraphics[width = 8.5cm]{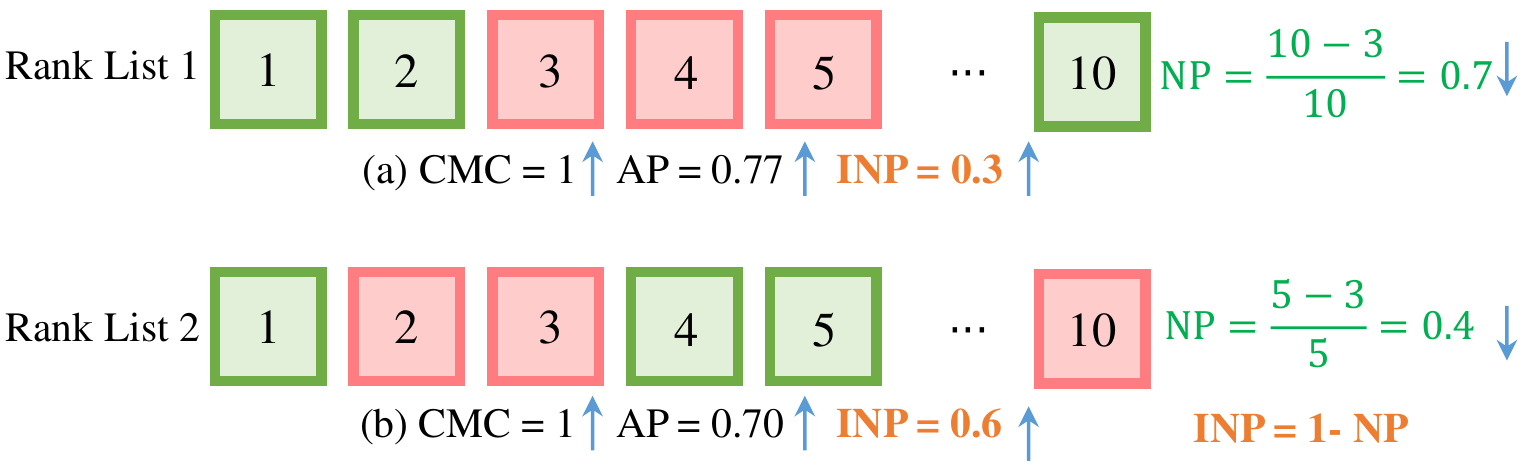}\\
  \vspace{-2mm}
  \caption{\blue{Difference between the widely used CMC, AP and the negative penalty (NP) measurements. True matching and false matching are bounded in green and red boxes, respectively. Assume that only three correct matches exist in the gallery, rank list 1 gets better AP, but gets much worse NP than rank list 2. The main reason is that rank list 1 contains too many false matchings before finding the hardest true matching. For consistency with CMC and mAP, we compute the inverse negative penalty (INP), \eg, INP = 1- NP. Larger INP means better  performance. }}\label{fig:ppmetric}
  \vspace{-2mm}
\end{figure}

\section{An Outlook: Re-ID in Next Era}\label{sec:future}
This section firstly presents a new evaluation metric in \S~\ref{sec:newmetric}, a strong baseline (in \S~\ref{sec:newbase}) for person Re-ID. It provides an important guidance for future Re-ID research. Finally, we discuss some under-investigated open issues in \S~\ref{sec:openissue}.
\subsection{mINP: A New Evaluation Metric for Re-ID}\label{sec:newmetric}
\blue{For a good Re-ID system, the target person should be retrieved as accurately as possible, \ie, all the correct matches should have low rank values. Considering that the target person should not be neglected in the top-ranked retrieved list, especially for multi-camera network, so as to accurately track the target. When the target person appears in the gallery set at multiple time stamps, the rank position of the hardest correct match determines the workload of the inspectors for further investigation. However, the currently widely used CMC and mAP metrics cannot evaluate this property, as shown in Fig.~\ref{fig:ppmetric}. With the same CMC, rank list 1 achieves a better AP than rank list 2, but it requires more efforts to find all the correct matches. To address this issue, we design a computationally efficient metric, namely a negative penalty (NP), which measures the penalty to find the hardest correct match}
\vspace{-2mm}
\begin{equation}\label{eq:np}
\mathrm{NP}_i = \frac{R_{i}^{hard}-|G_i|}{R_{i}^{hard}},
\end{equation}
where $R_{i}^{hard}$ indicates the rank position of the hardest match, and $|G_i|$ represents the total number of correct matches for query $i$. Naturally, a smaller NP represents better performance. For consistency with CMC and mAP, we prefer to use the inverse negative penalty (INP), an inverse operation of NP. Overall, the mean INP of all the queries is represented by
\vspace{-2mm}
\begin{equation}\label{eq:minp}
\mathrm{mINP} = \frac{1}{n} \sum\nolimits_i (1-\mathrm{NP}_i ) = \frac{1}{n} \sum\nolimits_i \frac{|G_i|}{R_{i}^{hard}}.
\end{equation}
\blue{The calculation of mINP is quite efficient and can be seamlessly integrated in the CMC/mAP calculating process. mINP avoids the domination of easy matches in the mAP/CMC evaluation. One limitation is that mINP value difference for large gallery size would be much smaller compared to small galleries. But it still can reflect the relative performance of a Re-ID model, providing a supplement to the widely-used CMC and mAP metrics.}

\begin{table}[t]\scriptsize
\centering
\caption{\label{tab:component}Comparison with the state-of-the-arts on single-modality image-based Re-ID. Rank-1 accuracy (\%), mAP (\%) and mINP (\%) are reported on two public datasets.}
 \vspace{-2mm}
\begin{tabular}{l|ccc|ccc}
  \hline
   & \multicolumn{3}{c|}{Market-1501 \cite{iccv15zheng}}&  \multicolumn{3}{c}{DukeMTMC \cite{iccv17duke}} \\
  Method             & R1    & mAP   &mINP   & R1    & mAP  & mINP  \\\hline
    BagTricks \cite{arxiv19trick} \tiny{CVPR19W}     & 94.5  &85.9 & 59.4 & 86.4  & 76.4  & 40.7   \\
    ABD-Net \cite{iccv19abdnet} \tiny{ICCV19} & \textbf{95.6} & \textbf{88.3} & \textbf{66.2} & 89.0   & 78.6 & 42.1   \\\hline
  B (ours)              & 94.2  & 85.4   & 58.3   & 86.1 & 76.1  & 40.3  \\
  B + Att \cite{cvpr18nonlocal} & 94.9 & 86.9 & 62.2 & 87.5 & 77.6 & 41.9 \\
    B + WRT                            & 94.6 & 86.8 & 61.9 & 87.1 & 77.0  & 41.4\\
      B + GeM \cite{pami18gem} & 94.4 & 86.3 & 60.1 & 87.3 & 77.3 & 41.9 \\
    B + WRT +  GeM             & 94.9 & 87.1 & 62.5 & 88.2 & 78.1 & 43.4\\
    AGW (Full)  & 95.1 & 87.8 & 65.0 & \textbf{89.0} & \textbf{79.6} & \textbf{45.7}\\\hline
 \end{tabular}
\end{table}
\subsection{A New Baseline for Single-/Cross-Modality Re-ID}\label{sec:newbase}
\black{According to the discussion in \S~\ref{sec:datasota}, we design a new \textit{AGW}\footnote{Details are in \url{https://github.com/mangye16/ReID-Survey} and comprehensive comparison is shown in the supplementary material.} baseline for person Re-ID, which achieves competitive performance on both single-modality (image and video) and cross-modality Re-ID tasks. Specifically, our new baseline is designed on top of BagTricks \cite{arxiv19trick}, and AGW contains the following three major improved components:}

(1) \textbf{Non-local Attention (Att) Block}. As discussed in \S~\ref{sec:datasota}, the attention scheme plays a crucial role in discriminative Re-ID model learning. We adopt the powerful non-local attention block \cite{cvpr18nonlocal} to obtain the weighted sum of the features at all positions, represented by
\vspace{-2mm}
\begin{equation}\label{eq:nonlocal}
\mathbf{z}_i = W_z * \phi(\mathbf{x}_i) + \mathbf{x}_i,
\end{equation}
where $W_z$ is a weight matrix to be learned, $\phi(\cdot)$ represents a non-local operation, and $+ \mathbf{x}_i$ formulates a residual learning strategy. Details can be found in \cite{cvpr18nonlocal}. We adopt the default setting from \cite{cvpr18nonlocal} to insert the non-local attention block.

(2) \textbf{Generalized-mean (GeM) Pooling}. As a fine-grained instance retrieval, the widely-used max-pooling or average pooling cannot capture the domain-specific discriminative features. We adopt a learnable pooling layer, named \textit{generalized-mean (GeM) pooling}  \cite{pami18gem}, formulated by
\begin{equation}\label{eq:gem}
\mathbf{f}  =[f_1 \cdots f_k \cdots f_K]^T, f_k = (\frac{1}{|\mathcal{X}_k|}\sum\nolimits_{x_i \in \mathcal{X}_k } x_i^{p_k})^{\frac{1}{p_k}},
\end{equation}
\black{where $f_k$ represents the feature map, and $K$ is number of feature maps in the last layer. $\mathcal{X}_k$ is the set of $W \times H$ activations for feature map $k \in \{1,2,\cdots,K\}$. $p_k$ is a pooling hyper-parameter, which is learned in the back-propagation process \cite{pami18gem}. The above operation approximates max pooling when $p_k \rightarrow \infty$ and average pooling when $p_k = 1$.}

(3) \textbf{Weighted Regularization Triplet (WRT) loss}. In addition to the baseline identity loss with softmax cross-entropy, we integrate with another weighted regularized triplet loss,
\vspace{-3mm}
\begin{equation}\label{eq:wtrloss}
\mathcal{L}_{wrt}(i) =  \log( 1+ \exp(\sum\nolimits_{j} w_{ij}^p d^p_{ij} - \sum\nolimits_k w_{ik}^n d^n_{ik})).
\end{equation}
\begin{equation}\label{eq:weights}
w_{ij}^p = \frac{\exp{(d^p_{ij})}}{\sum\nolimits_{d_{ij}^p\in \mathcal{P}_i}\exp(d^p_{ij})}, w_{ik}^n = \frac{\exp{(-d^n_{ik})}}{\sum\nolimits_{d^n_{ik}\in \mathcal{N}_i}\exp(-d^n_{ik})},
\end{equation}
\black{where $(i,j,k)$ represents a hard triplet within each training batch. For anchor $i$, $\mathcal{P}_i$ is the corresponding positive set, and $\mathcal{N}_i$ is the negative set. $d^p_{ij}$/$d^n_{ik}$ represents the pairwise distance of a positive/negative sample pair. The above weighted regularization inherits the advantage of relative distance optimization between positive and negative pairs, but it avoids introducing any additional margin parameters. Our weighting strategy is similar to \cite{cvpr19multisim}, but our solution does not introduce additional hyper-parameters.}

\begin{figure}[t]
  \centering
  \includegraphics[width = 9cm]{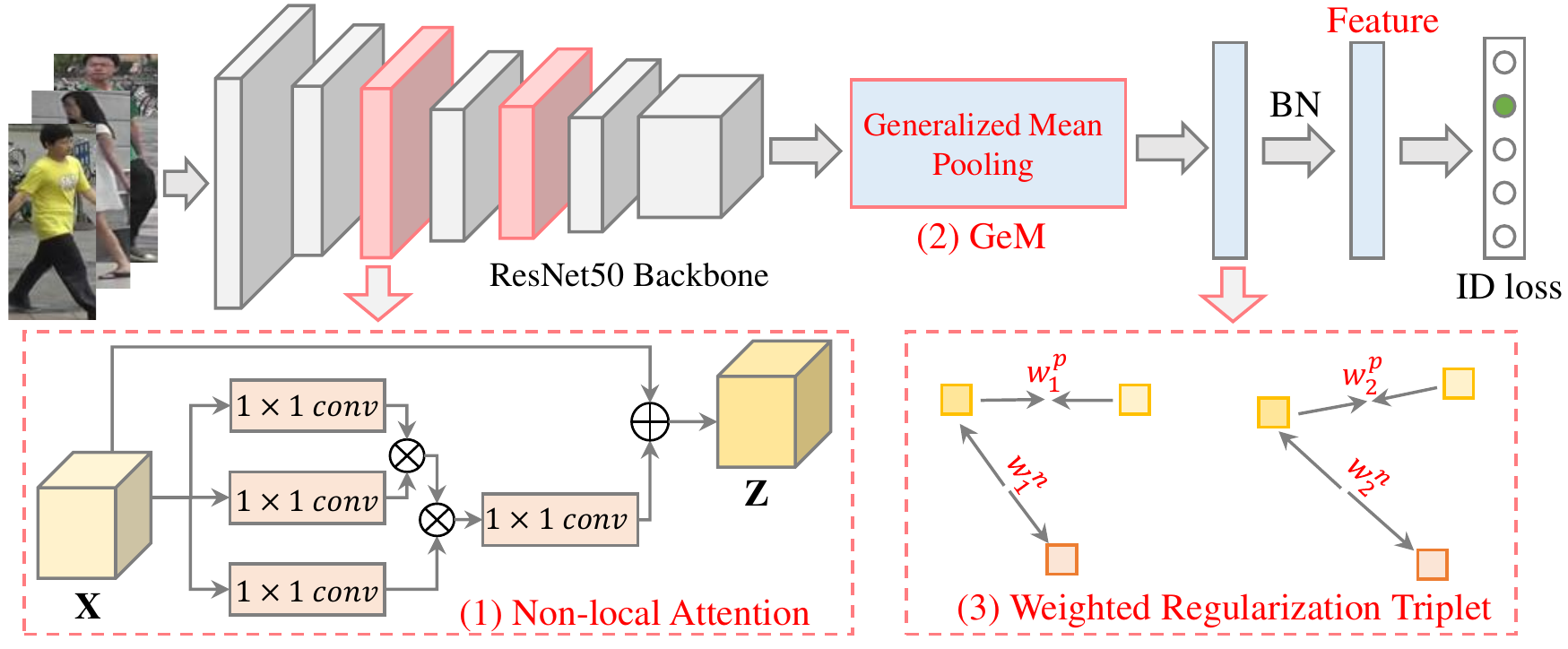}\\
  \vspace{-2mm}
  \caption{The framework of the proposed AGW baseline using the widely used ResNet50 \cite{cvpr16resnet} as the backbone network. }\label{fig:framework}
  \vspace{-2mm}
\end{figure}

\begin{table}[t]\scriptsize
\centering
\caption{\label{tab:agwcuhk}Comparison with the state-of-the-arts on two image Re-ID datasets, including CUHK03 and MSMT17. Rank-1 accuracy (\%), mAP (\%) and mINP (\%) are reported.}
 \vspace{-2mm}
\begin{tabular}{l|ccc|ccc}
  \hline

  & \multicolumn{3}{c|}{CUHK03 \cite{cvpr14cuhk}}&  \multicolumn{3}{c}{MSMT17 \cite{cvpr18msmt}}\\
      Method           & R1    & mAP  & mINP     & R1    & mAP   & mINP  \\\hline
  BagTricks \cite{arxiv19trick} \tiny{CVPR19W}      & 58.0 & 56.6 & 43.8 & 63.4  & 45.1 &   12.4  \\
  AGW (Full) & \textbf{63.6} & \textbf{62.0} & \textbf{50.3} & \textbf{68.3} & \textbf{49.3} & \textbf{14.7} \\\hline
 \end{tabular}
   \vspace{-2mm}
\end{table}

\begin{table}[t]\scriptsize
\centering
\caption{\label{tab:agwvid}Comparison with the state-of-the-arts on four video-based Re-ID datasets, including MARS \cite{eccv16mars}, DukeVideo \cite{cvpr18wu}, PRID2011 \cite{prid2011} and iLIDS-VID \cite{eccv14video}. Rank-1 accuracy (\%), mAP (\%) and mINP (\%) are reported.}
 \vspace{-2mm}
\begin{tabular}{l|ccc|ccc}
  \hline
  & \multicolumn{3}{c|}{MARS \cite{eccv16mars}}&  \multicolumn{3}{c}{DukeVideo \cite{cvpr18wu}}\\
      Method           & R1    & mAP  & mINP     & R1    & mAP   & mINP  \\\hline
  BagTricks \cite{arxiv19trick} \tiny{CVPR19W}      & 85.8 & 81.6 & 62.0 & 92.6  & 92.4 &   88.3  \\
CoSeg \cite{iccv19coseg} \tiny{ICCV19} & 84.9 & 79.9 & 57.8 & 95.4 &   94.1 & 89.8  \\
  AGW (Ours)  & 87.0 & 82.2 & 62.8 & 94.6 & 93.4 & 89.2 \\
    AGW$_+$ (Ours) & \textbf{87.6} & \textbf{83.0} & \textbf{63.9} & \textbf{95.4} & \textbf{94.9} & \textbf{91.9} \\\hline\hline
    & \multicolumn{3}{c|}{ PRID2011 \cite{prid2011}}&  \multicolumn{3}{c}{iLIDS-VID \cite{eccv14video}}\\
      Method           & R1    & R5  & mINP     & R1    & R5   & mINP  \\\hline
  BagTricks \cite{arxiv19trick} \tiny{CVPR19W}      & 84.3 & 93.3 &  88.5& 74.0  & 93.3 &   82.2 \\
  AGW (Ours) & 87.8 & 96.6 & 91.7 & 78.0 & 97.0 & 85.5 \\
  AGW$_+$ (Ours) & \textbf{94.4} &  \textbf{98.4} &  \textbf{95.4} &  \textbf{83.2} &  \textbf{98.3} &  \textbf{89.0} \\\hline
 \end{tabular}
   \vspace{-2mm}
\end{table}

\begin{table}[t]\scriptsize
\centering
\caption{\label{tab:agwpsid}Comparison with the state-of-the-arts on two partial Re-ID datasets, including Partial-REID and Partial-iLIDS. Rank-1, -3 accuracy (\%) and mINP (\%) are reported.}
 \vspace{-2mm}
\begin{tabular}{p{2.8cm}|p{0.5cm}<{\centering}p{0.5cm}<{\centering}p{0.65cm}<{\centering}|p{0.5cm}<{\centering}p{0.5cm}<{\centering}p{0.65cm}<{\centering}}
  \hline
   \multirow{2}{*}{Method} & \multicolumn{3}{c|}{Partial-REID}&  \multicolumn{3}{c}{Partial-iLIDS}\\
    \cline{2-7}
     &R1    &R3  & mINP     & R1    & R3   & mINP  \\
   \hline
  DSR \cite{cvpr18partial} \tiny{CVPR18} &50.7 &70.0 &- &58.8 &67.2 &- \\
  SFR \cite{arxiv2018partial} \tiny{ArXiv18} &56.9 &78.5 &- &63.9 &74.8 &- \\
  VPM \cite{cvpr19partial} \tiny{CVPR19}  &67.7 &\textbf{81.9} &- &\textbf{67.2} &76.5 &- \\
  BagTricks \cite{arxiv19trick} \tiny{CVPR19W}      &62.0 &74.0 &45.4 &58.8  &73.9 &68.7  \\
  \hline
  AGW  &\textbf{69.7} &80.0 &\textbf{56.7} &64.7 &\textbf{79.8} &\textbf{73.3} \\\hline
 \end{tabular}
   \vspace{-2mm}
\end{table}
\black{The overall framework of AGW is shown in Fig~\ref{fig:framework}. Other components are exactly the same as \cite{arxiv19trick}. In the testing phase, the output of BN layer is adopted as the feature representation for Re-ID. The implementation details and more experimental results are in the supplementary material.}

\black{\textbf{Results on Single-modality Image Re-ID.}
We first evaluate each component on two image-based datasets (Market-1501 and DukeMTMC) in Table \ref{tab:component}. We also list two state-of-the-art methods, BagTricks \cite{arxiv19trick} and ABD-Net \cite{iccv19abdnet}. We report the results on CUHK03 and MSMT17 datasets in Table \ref{tab:agwcuhk}. We obtain the following two observations:}

\black{1) All the components consistently contribute the accuracy gain, and AGW performs much better than the original BagTricks under various metrics. AGW provides a strong baseline for future improvements. We have also tried to incorporate part-level feature learning \cite{eccv18pcb}, but extensive experiments show that it does not improve the performance. How to aggregate  part-level feature learning with AGW needs further study in the future. 2) Compared to the current state-of-the-art, ABD-Net \cite{iccv19abdnet}, AGW performs favorably in most cases. In particular, we achieve much higher mINP on DukeMTMC dataset, 45.7\% \vs 42.1\%. This demonstrates that AGW requires less effort to find all the correct matches, verifying the ability of mINP.}

\blue{\textbf{Results on Single-modality Video Re-ID.}
We also evaluate the proposed AGW on four widely used single modality video-based datasets ( MARS \cite{eccv16mars}, DukeVideo \cite{cvpr18wu}, PRID2011 \cite{prid2011} and iLIDS-VID \cite{eccv14video}, as shown in Table \ref{tab:agwvid}. We also compare two state-of-the-art methods, BagTricks \cite{arxiv19trick} and Co-Seg \cite{iccv19coseg}. For video data, we develop a variant (AGW$_+$) to capture the temporal information with frame-level average pooling for sequence representation. Meanwhile, constraint random sampling strategy \cite{cvpr18diversityatt} is applied for training. Compared to Co-Seg \cite{iccv19coseg}, our AGW$_+$ obtains better Rank-1, mAP and mINP in most cases.}
\begin{table}[t]\scriptsize
\centering
\caption{\label{tab:crossreid}Comparison with the state-of-the-arts on cross-modality visible-infrared Re-ID. Rank-1 accuracy (\%), mAP (\%) and mINP (\%) are reported on two public datasets.}
 \vspace{-2mm}
\begin{tabular}{l|cc|cc|cc}
  \hline
   & \multicolumn{2}{c|}{RegDB \cite{sensors17}}&  \multicolumn{4}{c}{SYSU-MM01 \cite{iccv17cross}} \\
    & \multicolumn{2}{c|}{\emph{Visible-Thermal}}&  \multicolumn{2}{c}{\emph{All Search}} & \multicolumn{2}{c}{\emph{Indoor Search}} \\ \hline
  Method             & R1    & mAP   &  R1    & mAP    & R1    & mAP    \\\hline
  Zero-Pad  \cite{iccv17cross}  \tiny{ICCV17} &17.75      & 18.90  & 14.8 & 15.95 & 20.58 & 26.92    \\
  HCML \cite{aaai18vtreid}  \tiny{AAAI18} & 24.44    &20.08   & 14.32 &16.16 & 24.52 & 30.08     \\
  eBDTR \cite{tifs19vtreid}   \tiny{TIFS19}      & 34.62  & 33.46 & 27.82 & 28.42 & 32.46 &42.46 \\
  HSME \cite{aaai19hsme} \tiny{AAAI19} & 50.85   & 47.00   & 20.68 &23.12 & - & -\\
  D$^2$RL \cite{cvpr19ivreid} \tiny{CVPR19}  & 43.4   & 44.1 & 28.9 & 29.2 & - & -\\

  AlignG \cite{iccv19ivreid} \tiny{ICCV19}& 57.9  &  53.6 & 42.4 & 40.7 & 45.9 & 54.3\\
  Hi-CMD \cite{cvpr20hicmd} \tiny{CVPR20} & 70.93 & 66.04 &34.9 & 35.9 \\\hline
 \multirow{2}{*}{AGW (Ours)} & \textbf{70.05}    & \textbf{66.37}  & \textbf{47.50} & \textbf{47.65} & \textbf{54.17}& \textbf{62.97}\\
   & \multicolumn{2}{c|}{mINP = 50.19 } & \multicolumn{2}{c|}{mINP =35.30 }  & \multicolumn{2}{c}{mINP = 59.23}\\\hline
 \end{tabular}
 \vspace{-5mm}
\end{table}

\black{\textbf{Results on Partial Re-ID.}
We also test the performance of AGW on two partial Re-ID datasets, as shown in Table \ref{tab:agwpsid}. The experimental setting are from DSR \cite{cvpr18partial}. We also achieve comparable performance with the state-of-the-art VPM method \cite{cvpr19partial}. This experiment further demonstrates the superiority of AGW for the open-world partial Re-ID task. Meanwhile, the mINP also shows the applicability for this open-world Re-ID problem.}

\black{\textbf{Results on Cross-modality Re-ID.}
We also test the performance of AGW using a two-stream architecture on the cross-modality visible-infrared Re-ID task. The comparison with the current state-of-the-arts on two datasets is shown in Table \ref{tab:crossreid}. We follow the settings in AlignG \cite{iccv19ivreid} to perform the experiments. Results show that AGW achieves much higher accuracy than existing cross-modality Re-ID models, verifying the effectiveness for the open-world Re-ID task.}

\subsection{Under-Investigated Open Issues}\label{sec:openissue}
\black{We discuss the open-issues from five different aspects according to the five steps in \S \ref{sec:intro}, including uncontrollable data collection, human annotation minimization, domain-specific/generalizable architecture design, dynamic model updating and efficient model deployment.}

\subsubsection{Uncontrollable Data Collection}
\black{Most existing Re-ID works evaluate their method on a well-defined data collection environment. However, the data collection in real complex environment is uncontrollable. The data might be captured from unpredictable modality, modality combinations, or even cloth changing data \cite{pami19cloth}.}

\blue{\textbf{Multi-Heterogeneous Data.} In real applications, the Re-ID data might be captured from multiple heterogeneous modalities, \ie, the resolutions of person images vary a lot \cite{ijcai18resolution}, both the query and gallery sets may contain different modalities (visible, thermal \cite{iccv17cross}, depth \cite{tip17rgbd} or text description \cite{iccv17nlp}). This results in a challenging multiple heterogeneous person Re-ID. A good person Re-ID system would be able to automatically handle the changing resolutions, different modalities, various environments and multiple domains. Future work with broad generalizability is expected, evaluating their method for different Re-ID tasks.}

\black{\textbf{Cloth-Changing Data.} In practical surveillance system, it is very likely to contain a large number of target persons with changing clothes. A cloth-Clothing Change Aware
Network (CCAN) \cite{cvprw18cloth} addresses this issue by separately extracting the face and body context representation, and similar idea is applied in \cite{arxiv20cloth}. Yang \textit{et al.} \cite{pami19cloth} present a spatial polar transformation (SPT) to learn cross-cloth invariant representation. However, they still rely heavily on the face and body appearance, which might be unavailable and unstable in real scenarios. It would be interesting to further explore the possibility of other discriminative cues (e.g., gait, shape) to address the cloth-changing issue.}

\subsubsection{Human Annotation Minimization}
\black{Besides the unsupervised learning, active learning or human interaction \cite{eccv16temporal,icip15active,iccv19humanloop,eccv16humanloop} provides another possible solution to alleviate the reliance on human annotation.}

\black{\textbf{Active Learning.}
Incorporating human interaction, labels are easily provided for newly arriving data and the model can be subsequently updated \cite{icip15active,eccv16temporal}. A pairwise subset selection framework \cite{cvpr18labeling} minimizes human labeling effort by firstly constructing an edge-weighted complete $k$-partite graph and then solving it as a triangle free subgraph maximization problem. Along this line, a deep reinforcement active learning method \cite{iccv19humanloop} iteratively refines the learning policy and trains a Re-ID network with human-in-the-loop supervision. For video data, an interpretable reinforcement learning method with sequential decision making \cite{cvpr18decisionmaking} is designed. The active learning is crucial in practical Re-ID system design, but it has received less attention in the research community. Additionally, the newly arriving identities is extremely challenging, even for human. Efficient human in-the-loop active learning is expected in the future.}

\blue{\textbf{Learning for Virtual Data.}
This provides an alternative for minimizing the human annotation.
A synthetic dataset is collected in \cite{mm20synth} for training, and they achieve competitive performance on real-world datasets when trained on this synthesized dataset.
Bak \textit{et al.} \cite{eccv18bak} generate a new synthetic dataset with different illumination conditions to model realistic indoor and outdoor lighting.
A large-scale synthetic PersonX dataset is collected in \cite{cvpr19viewpoint} to systematically study the effect of viewpoint for a person Re-ID system. Recently, the 3D person images are also studied in \cite{arxiv203d}, generating the 3D body structure from 2D images. However, how to bridge the gap between synthesized images and real-world datasets remains challenging.}

\subsubsection{Domain-Specific/Generalizable Architecture Design}
\black{\textbf{Re-ID Specific Architecture.} Existing Re-ID methods usually adopt architectures designed for image classification as the backbone. Some methods modify the architecture to achieve better Re-ID features \cite{iccv17svd,arxiv19trick}. Very recently, researchers have started to design domain specific architectures, \eg, OSNet with omni-scale feature learning \cite{iccv19osnet}. It detects the small-scale discriminative features at a certain scale. OSNet is extremely lightweight and achieves competitive performance. With the advancement of automatic neural architecture search (\eg, Auto-ReID \cite{iccv19autoreid}), more domain-specific powerful architectures are expected to address the task-specific Re-ID challenges. Limited training samples in Re-ID also increase the difficulty in architecture design.}

\blue{\textbf{Domain Generalizable Re-ID.} It is well recognized that there is a large domain gap between different datsets \cite{icpr14deepmetric,eccv20liao}. Most existing methods adopt domain adaptation for cross-dataset training. A more practical solution would be learning a domain generalized model with a number of source datasets, such that the learned model can be generalized to new unseen datasets for discriminative Re-ID without additional training \cite{cvpr19general}. Hu \textit{et al.} \cite{accv14cross} studied the cross-dataset person Re-ID by introducing a part-level CNN framework. The Domain-Invariant Mapping Network (DIMN) \cite{cvpr19general} designs a meta-learning pipeline for domain generalizable Re-ID, learning a mapping between a person image and its identity classifier. The domain generalizability is crucial to deploy the learned Re-ID model under an unknown scenario.}

\subsubsection{Dynamic Model Updating}
\black{Fixed model is inappropriate for practical dynamically updated surveillance system. To alleviate this issue, dynamic model updating is imperative, either to a new domain/camera or adaptation with newly collected data.}

\black{\textbf{Model Adaptation to New Domain/Camera}. Model adaptation to a new domain has been widely studied in the literature as a domain adaptation problem \cite{eccv18bak,cvpr19adaptrans}. In practical dynamic camera network, a new camera may be temporarily inserted into an existing surveillance system. Model adaptation is crucial for continuous identification in a multi-camera network \cite{cviu17camera,eccv14camera}. To adapt a learned model to a new camera, a transitive inference algorithm \cite{cvpr17dynamicopen} is designed to exploit the best source camera model based on a geodesic flow kernel. However, it is still challenging when the newly collected data by the new camera has totally different distributions. In addition, the privacy and efficiency issue \cite{arxiv20fedreid} also need further consideration.}

\black{\textbf{Model Updating with Newly Arriving Data.} With the newly collected data, it is impractical to training the previously learned model from the scratch \cite{eccv16temporal}. An incremental learning approach together with human interaction is designed in \cite{eccv16temporal}. For deeply learned model, an addition using covariance loss \cite{arxiv18incremental} is integrated in the overall learning function. However, this problem is not well studied since the deep model training require large amount of training data. Besides, the unknown new identities in the newly arriving data is hard to be identified for the model updating. }

\subsubsection{Efficient Model Deployment}
\black{It is important to design efficient and adaptive models to address scalability issue for practical model deployment.}

\blue{\textbf{Fast Re-ID.}
For fast retrieval, hashing has been extensively studied to boost the searching speed, approximating the nearest neighbor search \cite{tip17hash}. Cross-camera Semantic Binary Transformation (CSBT) \cite{cvpr17fast} transforms the original high-dimensional feature representations into compact low-dimensional identity-preserving binary codes. A Coarse-to-Fine (CtF) hashing code search strategy is developed in \cite{eccv20faster}, complementarily using short and long codes. However, the domain-specific hashing still needs further study.}

\blue{\textbf{Lightweight Model.}
Another direction for addressing the scalability issue is to design a lightweight Re-ID model. Modifying the network architecture to achieve a lightweight model is investigated in \cite{cvpr18hann,iccv19autoreid,iccv19osnet}. Model distillation is another approach, \eg, a multi-teacher adaptive similarity distillation framework is proposed in \cite{cvpr19distillreid}, which learns a user-specified lightweight student model from multiple teacher models, without access to source domain data.}

\black{\textbf{Resource Aware Re-ID.}
Adaptively adjusting the model according to the hardware configurations also provides a solution to handle the scalability issue. Deep Anytime Re-ID (DaRe) \cite{cvpr18multires} employs a simple distance based routing strategy to adaptively adjust the model, fitting to hardware devices with different computational resources.}


\section{Concluding Remarks}\label{sec:conclusion}
This paper presents a comprehensive survey with in-depth analysis from a both closed-world and open-world perspectives. We first introduce the widely studied person Re-ID under the closed-world setting from three aspects: feature representation learning, deep metric learning and ranking optimization. With powerful deep learning, the closed-world person Re-ID has achieved performance saturation on several datasets. Correspondingly, the open-world setting has recently gained increasing attention, with efforts to address various practical challenges. We also design a new AGW baseline, which achieves competitive performance on four Re-ID tasks under various metrics. It provides a strong baseline for future improvements. This survey also introduces a new evaluation metric to measure the cost of finding all the correct matches. We believe this survey will provide important guidance for future Re-ID research.


\bibliographystyle{IEEEtran}
\bibliography{reidbib}


\vspace{3mm}
\noindent\textbf{\large Supplemental Materials:}

\setcounter{equation}{0}
\setcounter{figure}{0}
\setcounter{table}{0}
\setcounter{page}{1}
\makeatletter
\renewcommand{\theequation}{R\arabic{equation}}
\renewcommand{\thefigure}{R\arabic{figure}}

\renewcommand{\thefigure}{R\arabic{figure}}
\begin{figure*}[t]
  \centering
  \includegraphics[height = 6cm]{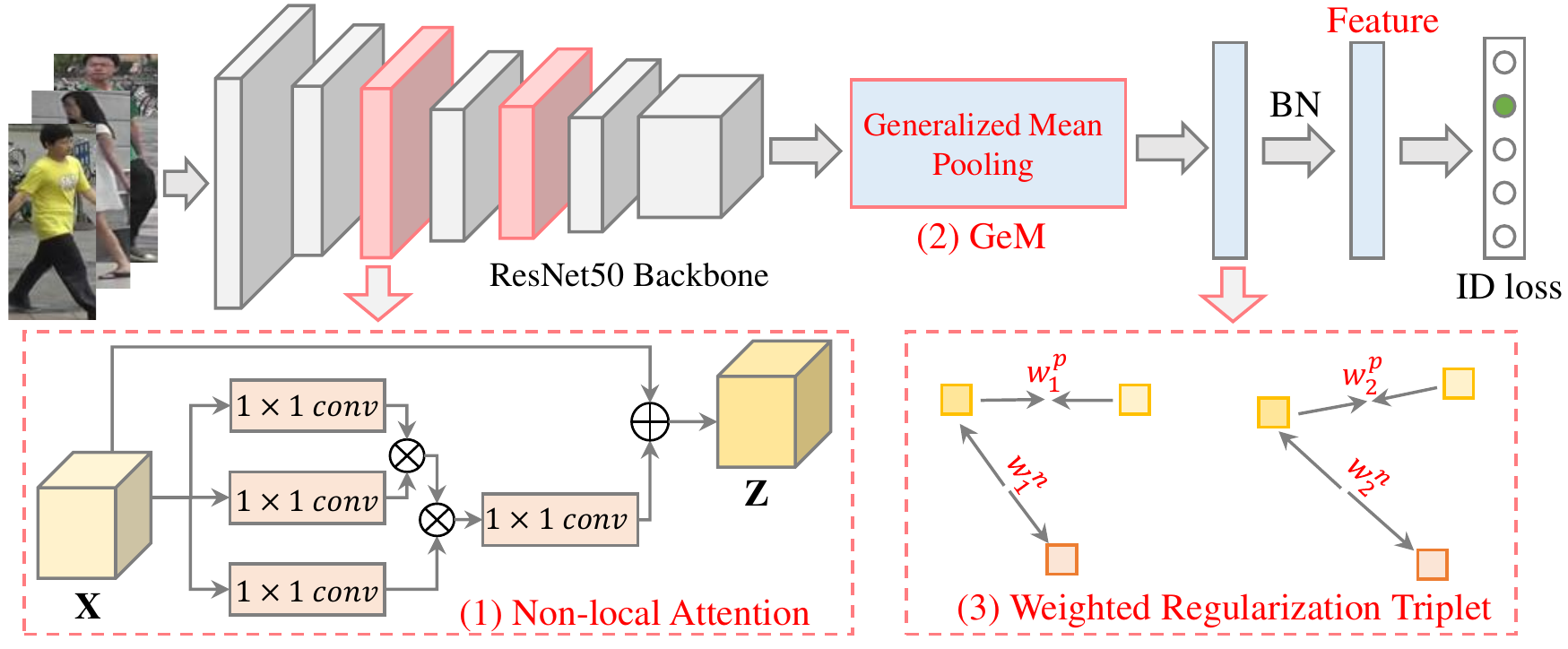}\\
  \caption{The framework of the proposed AGW baseline for single-modality image-based Re-ID. }\label{fig:sframework}
\end{figure*}
This supplementary material accompanies our main manuscript with the implementation details and more comprehensive experiments. We first present the experiments on two single-modality closed-world Re-ID tasks, including image-based Re-ID on four datasets in Section A and video-based Re-ID on four datasets in Section B. Then we introduce the comprehensive comparison on two open-world Re-ID tasks, including visible-infrared cross-modality Re-ID on two datasets in Section C and partial Re-ID on two datasets in Section D. In addition, a structure overview for our survey is finally summarized.

\subsection*{A. Experiments on Single-modality Image-based Re-ID}\label{secs:single}
\textbf{Architecture Design.}
The overall structure\footnote{\url{https://github.com/mangye16/ReID-Survey}} of our proposed AGW baseline for single-modality Re-ID is illustrated in \S~\ref{sec:future} (Fig.~\ref{fig:sframework}).
We adopt ResNet50 pre-trained on ImageNet as our backbone network and change the dimension of the fully connected layer to be consistent with the number of identities in the training dataset.
The stride of the last spatial down-sampling operation in the backbone network is changed from 2 to 1.
Consequently, the spatial size of the output feature map is changed from $8 \times 4$ to $16 \times 8$, when feeding an image of resolution $256 \times 128$ as input.
In our method, we replace the Global Average Pooling in the original ResNet50 with the Generalized-mean (GeM) pooling.
The pooling hyper parameter $p_k$ for generalized-mean pooling is initialized as 3.0.
A BatchNorm layer, named BNNeck is plugged between the GeM pooling layer and the fully connected layer.
The output of the GeM pooling layer is adopted for computing center loss and triplet loss in the training stage, while the feature after BNNeck is used for computing distance between pedestrian images during testing inference stage.

\textbf{Non-local Attention.}
The ResNet contains 4 residual stages, i.e. $conv2\_x$, $conv3\_x$, $conv4\_x$ and $conv5\_x$, each containing stacks of bottleneck residual blocks.
We inserted five non-local blocks after $conv3\_3$, $conv3\_4$, $conv4\_4$, $conv4\_5$ and $conv4\_6$ respectively.
We adopt the Dot Product version of non-local block with a bottleneck of 512 channels in our experiment.
For each non-local block, a BatchNorm layer is added right after the last linear layer that represents $W_z$.
The affine parameter of this BatchNorm layer is initialized as zeros to ensure that the non-local block can be inserted into any pre-trained
networks while maintaining its initial behavior.

\textbf{Training Strategy.}
In the training stage, we randomly sample 16 identities and 4 images for each identity to form a mini-batch of size 64.
Each image is resized into $256 \times 128$ pixels, padding 10 pixels with zero values, and then randomly cropped into $256 \times 128$ pixels.
Random horizontally flipping and random erasing with 0.5 probability respectively are also adopted for data augmentation.
Specifically, random erasing augmentation \cite{arxiv17randomerase} randomly selects a rectangle region with area ratio $r_e$ to the whole image, and erase its pixels with the mean value of the image.
Besides, the aspect ratio of this region is randomly initialized between $r_1$ and $r_2$.
In our method, we set the above hyper-parameter as $0.02<r_e<0.4$, $r1=0.3$ and $r2=3.33$.
At last, we normalize the RGB channels of each image with mean 0.485, 0.456, 0.406 and stand deviation 0.229, 0.224, 0.225, respectively, which are the same with settings in \cite{arxiv19trick}.

\renewcommand{\thetable}{R\arabic{table}}
\begin{table*}[t]\scriptsize
\centering
\caption{\label{tab:sp_agwvid}Comparison with the state-of-the-arts on four video-based Re-ID datasets, including MARS, DukeVideo, PRID2011 and iLIDS-VID. Rank-1, -5, -10 accuracy (\%), mAP (\%) and mINP (\%) are reported.}
\vspace{-2mm}
\begin{tabular}{p{1.65cm}|p{1.3cm}|p{0.4cm}<{\centering}p{0.4cm}<{\centering}p{0.45cm}<{\centering}p{0.65cm}<{\centering}|p{0.4cm}<{\centering}p{0.4cm}<{\centering}p{0.45cm}<{\centering}p{0.65cm}<{\centering}|p{0.4cm}<{\centering}p{0.4cm}<{\centering}p{0.45cm}<{\centering}p{0.65cm}<{\centering}|p{0.4cm}<{\centering}p{0.4cm}<{\centering}p{0.45cm}<{\centering}p{0.65cm}<{\centering}}
  \hline
  \multirow{2}{*}{Method} &\multirow{2}{*}{Venue}  &\multicolumn{4}{c|}{MARS}  & \multicolumn{4}{c|}{DukeVideo} &\multicolumn{4}{c|}{PRID2011} &\multicolumn{4}{c}{iLIDS-VID} \\
  \cline{3-18}
    &  &R1   &R5   &mAP &mINP   &R1   &R5   &mAP &mINP  &R1   &R5   &R20 &mINP &R1   &R5   &R20 &mINP  \\
  \hline
  ETAP~\cite{cvpr18wu} &\footnotesize{CVPR18} &80.7  &92.0  &67.3 &-  &83.6 &94.5 &78.3 &- &- &- &- &- &- &- &- &- \\
  DRSA~\cite{cvpr18diversityatt} &\footnotesize{CVPR18} &82.3  &-  &65.8 &-  &- &- &- &- &93.2 &- &- &- &80.2 &- &- &- \\
  Snippet~\cite{cvpr18snippet} &\footnotesize{CVPR18}  &86.3  &94.7  &76.1 &-  &- &- &- &- &93.0 &99.3 &- &- &85.4 &96.7 &- &- \\
  STA~\cite{aaai19sta}  &\footnotesize{AAAI18}  &86.3  &95.7  &80.8 &-  &96.2 &99.3 &94.9 &- &- &- &- &- &- &- &- &- \\
  VRSTC~\cite{cvpr19occlusion} &\footnotesize{CVPR19} &88.5  &96.5  &82.3 &-  &95.0 &99.1  &93.5 &- &- &- &- &- &83.4 &95.5 &99.5 &- \\
  ADFD~\cite{cvpr19videoatt} &\footnotesize{CVPR19} &87.0  &95.4  &78.2 &-  &- &-  &- &- &93.9 &99.5 &100 &- &86.3 &97.4 &99.7 &- \\
  GLTR~\cite{iccv19longshort} &\footnotesize{ICCV19} &87.0  &95.7  &78.4 &-  &96.2 &99.3  &93.7 &- &95.5 &100.0 &- &- &86.0 &98.0 &- &- \\
  CoSeg~\cite{iccv19coseg} &\footnotesize{ICCV19}  &84.9  &95.5  &79.9  &57.8  &95.4 &99.3  &94.1 &89.8 &- &- &- &- &79.6 &95.3 &99.3 &- \\
  BagTricks~\cite{arxiv19trick} &\footnotesize{CVPR19W}   &85.8  &95.2  &81.6 &62.0  &92.6   &98.9   &92.4 &88.3  &84.3 &93.3 &98.0 &88.5  &74.0 &93.3 &99.1 &82.2 \\
  \hline
  AGW &- &87.0 &95.7 &82.2 &62.8 &94.6 &99.1 &93.4	&89.2 &87.8 &96.6 &98.9 &91.7  &78.0 &97.0 &99.5 &85.5 \\
  AGW$_+$ &-  &87.6 &85.8 &83.0 &63.9 &95.4 &99.3 &94.9 &91.9  &94.4 &98.4 &100 &95.4 &83.2 &98.3 &99.7 &89.0\\\hline
 \end{tabular}
\end{table*}

\textbf{Training Loss.}
In the training stage, three types of loss are combined for optimization, including identity classification loss ($\mathcal{L}_{id}$), center loss  ($\mathcal{L}_{ct}$) and our proposed weighted regularization triplet loss ($\mathcal{L}_{wrt}$).
\begin{equation}
\mathcal{L}_{total} = \mathcal{L}_{id} + \beta_1\mathcal{L}_{ct} + \beta_2\mathcal{L}_{wrt}.
\end{equation}
The balanced weight of the center loss ($\beta_1$) is set to 0.0005 and the one ($\beta_2$) of the weighted regularized triplet loss is set to 1.0.
Label smoothing is adopted to improve the original identity classification loss, which encourages the model to be less confident during training and prevent overfitting for classification task.
Concretely, it changes the one-hot label as follow:
\begin{equation}
q_{i}=\left\{\begin{array}{ll}
{1-\frac{N-1}{N} \varepsilon} & {\text { if } i=y} \\
{\varepsilon / N} & {\text { otherwise }},
\end{array}\right.
\end{equation}
where $N$ is the total number of identities, $\epsilon$ is a small constant to reduce the confidence for the true identity label $y$ and $q_i$ is treated as a new classification target for training.
In our method, we set $\epsilon$ to be 0.1.

\textbf{Optimizer Setting.}
Adam optimizer with a weight decay $0.0005$ is adopted to train our model.
The initial learning rate is set as 0.00035 and is decreased by 0.1 at the 40th epoch and 70th epoch, respectively.
The model is trained for 120 epochs in total.
Besides, a warm-up learning rate scheme is also employed to improve the stability of training process and bootstrap the network for better performance. Specifically, in the first 10 epochs, the learning rate is linearly increased from $3.5 \times 10^{-5}$ to $3.5 \times 10^{-4}$.
The learning rate $lr(t)$ at epoch $t$ can be computed as:
\begin{equation}
\operatorname{lr}(t)=\left\{\begin{array}{ll}
{3.5 \times 10^{-5} \times \frac{t}{10}} & {\text { if } t \leq 10} \\
{3.5 \times 10^{-4}} & {\text { if } 10<t \leq 40} \\
{3.5 \times 10^{-5}} & {\text { if } 40<t \leq 70} \\
{3.5 \times 10^{-6}} & {\text { if } 70<t \leq 120}.
\end{array}\right.
\end{equation}

\begin{figure*}[t]
  \centering
  \includegraphics[height = 5.6cm]{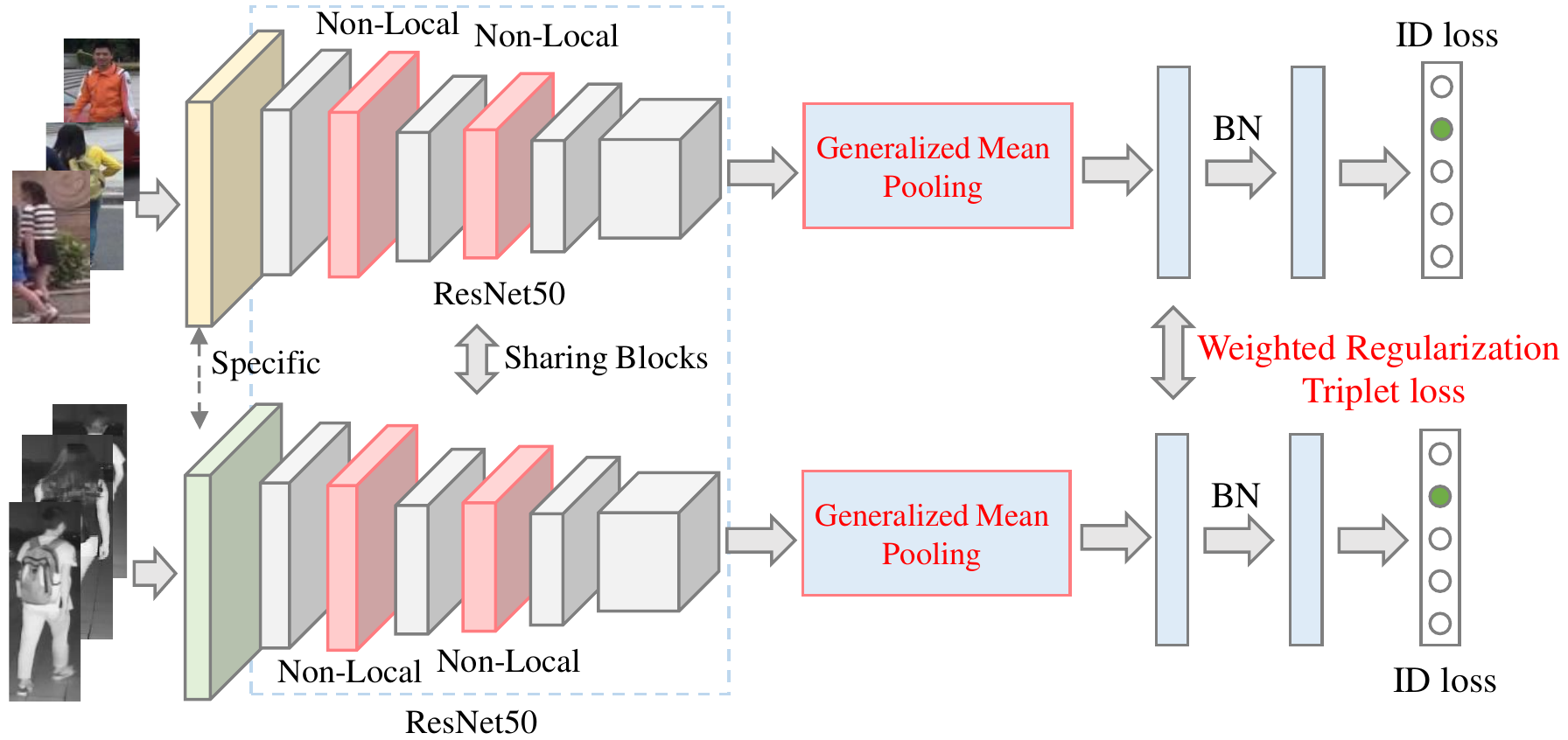}\\
  \caption{The framework of the proposed AGW baseline for cross-modality visible-infrared Re-ID. }\label{fig:crossframework}
\end{figure*}

 \renewcommand{\thetable}{R\arabic{table}}
\begin{table*}[t]\scriptsize
\centering
 \caption{\label{tab:sysu}Comparison with the state-of-the-arts on SYSU-MM01 dataset. Rank at $r$ accuracy (\%), mAP (\%)  and mINP (\%) are reported. (\textit{Single-shot query setting \cite{iccv17cross} for experiments)}. ``*" represents methods published after the paper submission.}
 \vspace{-2mm}
\begin{tabular}{p{2.2cm}|p{1.2cm}|p{1cm}<{\centering}p{1cm}<{\centering}p{1cm}<{\centering}|p{1cm}<{\centering}p{0.9cm}<{\centering}||p{1cm}<{\centering}p{1cm}<{\centering}p{1cm}<{\centering}|p{1.0cm}<{\centering}p{0.9cm}<{\centering}}
\hline
\multicolumn{2}{c|}{Settings} &\multicolumn{5}{c||}{\textit{All Search}}  & \multicolumn{5}{c}{\textit{Indoor Search}} \\\hline
Method   &Venue    & $r=1$     & $r=10$    & $r=20$ & mAP     & mINP       & $r=1$     & $r=10$    & $r=20$ & mAP  & mINP\\\hline
  One-stream \cite{iccv17cross} &\footnotesize{ICCV17}& 12.04 & 49.68   & 66.74       & 13.67 & - & 16.94 & 63.55   & 82.10     & 22.95    & - \\
  Two-stream \cite{iccv17cross} &\footnotesize{ICCV17}& 11.65 & 47.99  & 65.50       & 12.85  & -& 15.60 & 61.18   & 81.02     & 21.49   & -  \\
  Zero-Pad \cite{iccv17cross} &\footnotesize{ICCV17}&14.80 &54.12   &71.33       &15.95  & -& 20.58 & 68.38   & 85.79     & 26.92   & -  \\
  TONE \cite{aaai18vtreid} &\footnotesize{AAAI18}&12.52 & 50.72   &68.60      & 14.42      & -     & 20.82 & 68.86  & 84.46     & 26.38  & -    \\
  HCML \cite{aaai18vtreid}  &\footnotesize{AAAI18}&14.32 &53.16   &69.17  & 16.16  & - & 24.52  & 73.25  & 86.73     & 30.08   & - \\
  BDTR \cite{tifs19vtreid}  &\footnotesize{IJCAI18}& 27.32 & 66.96 & 81.07 & 27.32 & - & 31.92 & 77.18 & 89.28 & 41.86 & - \\
  eBDTR \cite{tifs19vtreid}  &\footnotesize{TIFS19}& 27.82 & 67.34 & 81.34 & 28.42 & - & 32.46 & 77.42 & 89.62 & 42.46& - \\
  HSME \cite{aaai19hsme} &\footnotesize{AAAI19}& 20.68 & 32.74 & 77.95  & 23.12 & -& - & -& -& -& - \\
  D$^2$RL \cite{cvpr19ivreid}  &\footnotesize{CVPR19}& 28.9 & 70.6  & 82.4 & 29.2 & -  & - & -& -& - & -\\
  MAC \cite{mm19mac}      &\footnotesize{MM19}& 33.26  & 79.04 & 90.09   & 36.22   & -    & 36.43   & 62.36  & 71.63 & 37.03 & -\\
  MSR \cite{tip19msr} &\footnotesize{TIP19}& 37.35 & 83.40 & 93.34 & 38.11 & - &  39.64  & 89.29   & 97.66   &  50.88 & - \\
  AlignGAN \cite{iccv19ivreid}&\footnotesize{ICCV19} & 42.4 & 85.0 & 93.7 & 40.7& - & 45.9 & 87.6 & 94.4 & 54.3 & - \\
    X-Modal$^{*}$ \cite{aaai20xmodal} & AAAI-20 & 49.9 &  89.8 & 96.0 & 50.7& - & -& - & -& -& - \\
  Hi-CMD$^{*}$ \cite{cvpr20hicmd} & CVPR20 & 34.9  &  77.6 & - & 35.9 & -& -  &- & -& -& - \\
      cm-SSFT$^{*}$  \cite{cvpr20ssft}\tnote{\dag} & CVPR20 &  47.7  & - & - & 54.1 & -  & -  & - & -  & - & - \\
  DDAG$^{*}$ \cite{eccv20ddag}     & ECCV20  & 54.75  & 90.39 & 95.81  & 53.02 & 39.62& 61.02  &94.06 & 98.41 & 67.98  & 62.61\\
 HAT$^{*}$ \cite{tifs20gray} & TIFS20 & 55.29  & 92.14 & 97.36 & 53.89 & - & 62.10 & 95.75    & 99.20 & 69.37 & - \\\hline
 AGW     & \centering{-} & 47.50  & 84.39 & 92.14 & 47.65 & 35.30  & 54.17  & 91.14 & 95.98 & 62.97 & 59.23\\\hline
 \end{tabular}
\end{table*}

 \renewcommand{\thetable}{R\arabic{table}}
\begin{table*}[t]\scriptsize
\centering
\caption{\label{tab:regdb}Comparison with the state-of-the-arts on RegDB dataset on both query settings. Rank at $r$ accuracy (\%), mAP (\%)  and mINP (\%) are reported. (\textit{Both the visible to thermal and thermal to visible query settings are evaluated.)} ``*" represents methods published after the paper submission.}
 \vspace{-2mm}
\begin{tabular}{p{2.0cm}|p{1.2cm}|p{1cm}<{\centering}p{1cm}<{\centering}p{1cm}<{\centering}|p{1cm}<{\centering}p{0.9cm}<{\centering}||p{1cm}<{\centering}p{1cm}<{\centering}p{1cm}<{\centering}|p{1.1cm}<{\centering}p{0.9cm}<{\centering}}
  \hline
  \multicolumn{2}{c|}{Settings} &\multicolumn{5}{c||}{\textit{Visible to Thermal}}  & \multicolumn{5}{c}{\textit{Thermal to Visible}} \\\hline
  Method  &Venue  & $r=1$   & $r=10 $   & $r=20$   & mAP & mINP   & $r=1$   & $r=10 $   & $r=20$   & mAP & mINP  \\\hline
  HCML \cite{aaai18vtreid} &\footnotesize{AAAI18}  & 24.44   &47.53  &  56.78 &20.08     & -   &   21.70   &45.02  &  55.58 & 22.24     & - \\
  Zero-Pad  \cite{iccv17cross} &\footnotesize{ICCV17}  &17.75 &34.21   &44.35     & 18.90   & -   & 16.63 & 34.68   &44.25     & 17.82  & - \\
  BDTR \cite{tifs19vtreid}  &\footnotesize{IJCAI18}    & 33.56  & 58.61   & 67.43     & 32.76  & - & 32.92  & 58.46   & 68.43    & 31.96 & -    \\
  eBDTR \cite{tifs19vtreid}  &\footnotesize{TIFS19}       & 34.62 & 58.96 & 68.72 & 33.46  & -  & 34.21 & 58.74 & 68.64 & 32.49  & -\\
  HSME \cite{aaai19hsme}&\footnotesize{AAAI19} & 50.85 & 73.36 & 81.66  & 47.00 & - &50.15 & 72.40 & 81.07  & 46.16  & -\\
  D$^2$RL \cite{cvpr19ivreid} &\footnotesize{CVPR19}& 43.4  & 66.1 & 76.3 & 44.1  & - & -& -& -& -& -\\
  MAC \cite{mm19mac}&\footnotesize{MM19} & 36.43   & 62.36  & 71.63 & 37.03  & - &36.20  & 61.68   & 70.99     & 36.63  & -\\
  MSR \cite{tip19msr} &\footnotesize{TIP19}& 48.43 & 70.32 & 79.95 & 48.67  & -  & -& -& -& -& -\\
  AlignGAN \cite{iccv19ivreid} &\footnotesize{ICCV19}& 57.9 & - & - &  53.6  & -& 56.3 & - & - &  53.4  & -\\
    XModal$^{*}$ \cite{aaai20xmodal} & AAAI20 & 62.21 & 83.13 & 91.72 & 60.18 & - & -& -& -  & -& -\\
     Hi-CMD$^{*}$ \cite{cvpr20hicmd} & CVPR20 & 70.93  & 86.39 & - &  66.04  & - & -& -& - & - & -\\
      cm-SSFT$^{*}$ \cite{cvpr20ssft}& CVPR20 &  72.3  & - & -  & 72.9 & - & 71.0  & - & -& 71.7 & - \\
  DDAG$^{*}$ \cite{eccv20ddag} & ECCV20 & 69.34    & 86.19 & 91.49 & 63.46 & 49.24 & 68.06 & 85.15  & 90.31 & 61.80 & 48.62\\
    HAT$^{*}$ \cite{tifs20gray} & TIFS20 & 71.83    & 87.16 & 92.16 & 67.56 &  - & 70.02  & 86.45  & 91.61 & 66.30 & -\\ \hline
   AGW  &- & 70.05   & 86.21 & 91.55 & 66.37 & 50.19 & 70.49   & 87.12 & 91.84 & 65.90 & 51.24\\\hline
 \end{tabular}
\end{table*}

\subsection*{B. Experiments on Video-based Re-ID}
\textbf{Implementation Details.}
We extend our proposed AGW baseline to a video-based Re-ID model by several minor changes to the backbone structure and training strategy of single-modality image-based Re-ID model.
The video-based AGW baseline takes a video sequence as input and extracts the frame-level feature vectors, which are then averaged to be a video-level feature vector before the BNNeck layer.
Besides, the video-based AGW baseline is trained for 400 epochs totally to better fit the video person Re-ID datasets.
The learning rate is decayed by 10 times every 100 epochs.
To form an input video sequence, we adopt the constraint random sampling strategy \cite{cvpr18diversityatt} to sample 4 frames as a summary for the original pedestrian tracklet.
The BagTricks~\cite{arxiv19trick} baseline is extended to a video-based Re-ID model in the same way as AGW baseline for fair comparison.
In addition, we also develop a variant of AGW baseline, termed as AGW$_+$, to model more abundant temporal information in a pedestrian tracklet.
AGW$_+$ baseline adopts the dense sampling strategy to form an input video sequence in the testing stage.
Dense sampling strategy takes all the frames in a pedestrian tracklet to form input video sequence, resulting better performance but higher computational cost.
To further improve the performance of AGW$_+$ baseline on video re-ID datasets, we also remove the warm-up learning rate strategy and add dropout operation before the linear classification layer.

\textbf{Detailed Comparison.}
In this section, we conduct the performance comparison between AGW baseline and other state-of-the-art video-based person Re-ID methods, including  ETAP~\cite{cvpr18wu}, DRSA~\cite{cvpr18diversityatt}, STA~\cite{aaai19sta} Snippet~\cite{cvpr18snippet},  VRSTC~\cite{cvpr19occlusion}, ADFD~\cite{cvpr19videoatt}, GLTR~\cite{iccv19longshort} and CoSeg~\cite{iccv19coseg}.
The comparison results on four video person Re-ID datasets (MARS, DukeVideo, PRID2011 and iLIDS-VID) are listed in Table~\ref{tab:sp_agwvid}.
As we can see, by simply taking video sequence as input and adopting average pooling to aggregate frame-level feature, our AGW baseline achieves competitive results on two large-scale video Re-ID dataset, MARS and DukeVideo.
Besides, AGW baseline also performs significantly better than BagTricks~\cite{arxiv19trick} baseline under multiple evaluation metrics.
By further modeling more temporal information and adjusting training strategy, AGW$_+$ baseline gains huge improvement and also achieves competitive results on both PRID2011 and iLIDS-VID datasets.
AGW$_+$ baseline outperforms most state-of-the-art methods on MARS, DukeVideo and PRID2011 datasets.
Most of these video-based person Re-ID methods achieve state-of-the-art performance by designing complicate temporal attention mechanism to exploit temporal dependency in pedestrian video.
We believe that our AGW baseline can help video Re-ID model achieve higher performance with properly designed mechanism to further exploit spatial and temporal dependency.

\subsection*{C. Experiments on Cross-modality Re-ID}
\textbf{Architecture Design.} We adopt a two-stream network structure as the backbone for cross-modality visible-infrared Re-ID\footnote{\url{https://github.com/mangye16/Cross-Modal-Re-ID-baseline}}. Compared to the one-stream architecture in single-modality person Re-ID (Fig.~\ref{fig:framework}), the major difference is that, \ie, the first block is specific for two modalities in order to capture modality-specific information, while the remaining blocks are shared to learn modality sharable features. Compared to the two-stream structure widely used in \cite{tifs19vtreid,mm19mac}, which only has one shared embedding layer, our design captures more sharable components. An illustration for cross-modality visible-infrared Re-ID is shown in Fig.~\ref{fig:crossframework}.

\textbf{Training Strategy.} At each training step, we random sample 8 identities from the whole dataset. Then 4 visible and 4 infrared images are randomly selected for each identity. Totally, each training batch contains 32 visible and 32 infrared images. This guarantees the informative hard triplet mining from both modalities, \ie, we directly select the hard positive and negative from both intra- and inter-modalities. This approximates the idea of bi-directional center-constrained top-ranking loss, handling the inter- and intra-modality variations simultaneously.

For fair comparison, we follow the settings in \cite{tifs19vtreid} exactly to conduct the image processing and data augmentation. For infrared images, we keep the original three channels, just like the visible RGB images. All the input images from both modalities are first resized to $288\times 144$, and random crop with zero-padding together with random horizontal flipping are adopted for data argumentation. The cropped image sizes are $256 \times 128$ for both modality. The image normalization are exactly following the single-modality setting.

\textbf{Training Loss.}
In the training stage, we combine with the identity classification loss ($\mathcal{L}_{id}$) and our proposed weighted regularization triplet loss ($\mathcal{L}_{wrt}$). The weight of combining the identity loss and weighted regularized triplet loss is set to 1, the same as the single-modality setting. The pooling parameter $p_k$ is set to 3. For stable training, we adopt the same identity classifier for two heterogeneous modalities, mining sharable information.

\textbf{Optimizer Setting.} We set the initial learning rate as 0.1 on both datasets, and decay it by 0.1 and 0.01 at 20 and 50 epochs, respectively. The total number of training epoch is 60. We also adopt a warm-up learning rate scheme. We adopt the stochastic gradient descent (SGD) optimizer for optimization, and the momentum parameter is set to 0.9. We have tried the same Adam optimizer (used in single-modality Re-ID) on cross-modality Re-ID task, but the performance is much lower than that of SGD optimizer by using a large learning rate. This is crucial since ImageNet initialization is adopted for the infrared images.

\textbf{Detailed Comparison}
This section conducts the comparison with the state-of-the-art cross-modality VI-ReID methods, including eBDTR \cite{tifs19vtreid}, HSME \cite{aaai19hsme}, D$^2$RL \cite{cvpr19ivreid}, MAC \cite{mm19mac}, MSR \cite{tip19msr} and AlignGAN \cite{iccv19ivreid}. These methods are published in the past two years. AlignGAN \cite{iccv19ivreid}, published in ICCV 2019, achieves the state-of-the-art performance by aligning the cross-modality representation at both the feature level and pixel level with GAN generated images.
The results on two datasets are shown in Tables \ref{tab:sysu} and \ref{tab:regdb}. We observe that the proposed AGW consistently outperforms the current state-of-the-art, without the time-consuming image generation process. For different query settings on RegDB dataset, our proposed baseline  generally keeps the same performance. Our proposed baseline has been widely used in many recently developed methods. We believe our new baseline will provide a good guidance to boost the cross-modality Re-ID.
\begin{figure*}[h]
\centering
  \includegraphics[width = 16cm]{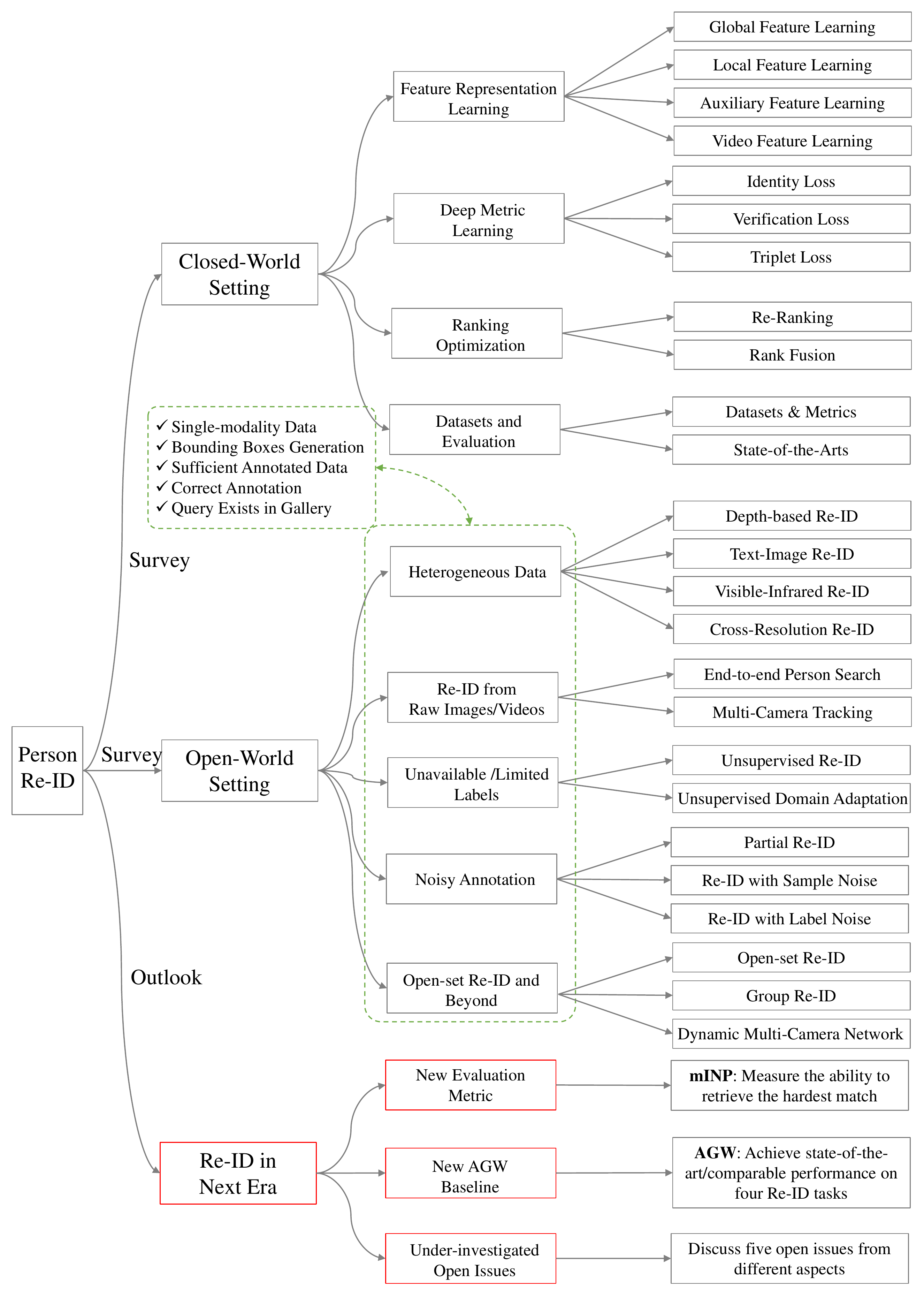}\\
  \caption{An overview of this survey. It contains three main components, including Closed-World Setting in Section \ref{sec:close}, Open-World Setting in Section \ref{sec:open} and an outlook of Re-ID in Next Era in Section \ref{sec:future}. }\label{fig:structure}
\end{figure*}

\subsection*{D. Experiments on Partial Re-ID}

\textbf{Implementation Details.}
We also evaluate the performance of our proposed AGW baseline on two commonly-used partial Re-ID datasets, Partial-REID and Partial-iLIDS.
The overall backbone structure and training strategy for partial Re-ID AGW baseline model are the same as the one for single-modality image-based Re-ID model.
Both Partial-REID and Partial-iLIDS datasets offer only query image set and gallery image set.
So we train AGW baseline model on the training set of Market-1501 dataset, then evaluate its performance on the testing set of two partial Re-ID datasets.
We adopt the same way to evaluate the performance of BagTricks \cite{arxiv19trick} baseline on these two partial Re-ID datasets for better comparison and analysis.

\textbf{Detailed Comparison.}
We compare the performance of AGW baseline with other state-of-the-art partial Re-ID methods, including DSR \cite{cvpr18partial}, SFR \cite{arxiv2018partial} and VPM \cite{cvpr19partial}.
All these methods are published in recent years.
The comparison results on both Partial-REID and Partial-iLIDS datasets are shown in Table~\ref{tab:agwpid}.
The VPM \cite{cvpr19partial} achieves a very high performance by perceiving the visibility of regions through self-supervision and extracting region-level features.
Considering only global features, our proposed AGW baseline still achieves competitive results compared to the current state-of-the-arts on both datasets.
Besides, AGW baseline brings significant improvement comparing to BagTricks \cite{arxiv19trick} under multiple evaluation metrics, demonstrating its effectiveness for partial Re-ID problem.

\renewcommand{\thetable}{R\arabic{table}}
\begin{table}[t]\scriptsize
\centering
\caption{\label{tab:agwpid}Comparison with the state-of-the-arts on two partial Re-ID datasets, including Partial-REID and Partial-iLIDS. Rank-1, -3 accuracy (\%) and mINP (\%) are reported.}
\vspace{-2mm}
\begin{tabular}{p{2.55cm}|p{0.5cm}<{\centering}p{0.5cm}<{\centering}p{0.65cm}<{\centering}|p{0.5cm}<{\centering}p{0.5cm}<{\centering}p{0.65cm}<{\centering}}
  \hline
   \multirow{2}{*}{Method} & \multicolumn{3}{c|}{Partial-REID}&  \multicolumn{3}{c}{Partial-iLIDS}\\
    \cline{2-7}
     &R1    &R3  & mINP     & R1    & R3   & mINP  \\
   \hline
  DSR \cite{cvpr18partial} \tiny{CVPR18} &50.7 &70.0 &- &58.8 &67.2 &- \\
  SFR \cite{arxiv2018partial} \tiny{ArXiv18} &56.9 &78.5 &- &63.9 &74.8 &- \\
  VPM \cite{cvpr19partial} \tiny{CVPR19}  &67.7 &\textbf{81.9} &- &\textbf{67.2} &76.5 &- \\
  BagTricks \cite{arxiv19trick} \tiny{CVPR19W}      &62.0 &74.0 &45.4 &58.8  &73.9 &68.7  \\
  \hline
  AGW  &\textbf{69.7} &80.0 &\textbf{56.7} &64.7 &\textbf{79.8} &\textbf{73.3} \\\hline
 \end{tabular}
\end{table}

\subsection*{E. Overview of This Survey}
The overview figure of this survey is shown in Fig.~\ref{fig:structure}. According to the five steps in developing a person Re-ID system, we conduct the survey from both closed-world and open-world settings. The closed-world setting is detailed in three different aspects: feature representation learning, deep metric learning and ranking optimization. We then summarize the datasets and SOTAs from both image- and video-based perspectives. For open-world person Re-ID, we summarize it into five aspects: including heterogeneous data, Re-ID from raw images/videos, unavailable/limited labels, noisy annotation and open-set Re-ID.

Following the summary, we present an outlook for future person Re-ID. We design a new evaluation metric (mINP) to evaluate the difficulty to find all the correct matches. By analyzing the advantages of existing Re-ID methods, we develop a strong AGW baseline for future developments, which achieves competitive performance on four Re-ID tasks. Finally, some under-investigated open issues are discussed. Our survey provides a comprehensive summarization of existing state-of-the-art in different sub-tasks. Meanwhile, the analysis of future directions is also presented for further development guidance.

\textbf{Acknowledgement.} The authors would like to thank the anonymous reviewers for providing valuable feedbacks to improve the quality of this survey. The authors also would like to thank the pioneer researchers in person re-identification and other related fields. This work is sponsored by CAAI-Huawei MindSpore Open Fund.
\end{document}